\documentclass{article}

 \usepackage[preprint]{neurips_2025}

\usepackage[utf8]{inputenc} 
\usepackage[T1]{fontenc}    
\usepackage{CJKutf8}
\usepackage{hyperref}       
\usepackage{url}            
\usepackage{booktabs}       
\usepackage{amsfonts}       
\usepackage{amsmath}
\usepackage{nicefrac}       
\usepackage{microtype}      
\usepackage{xcolor}         
\definecolor{darkgreen}{rgb}{0.0,0.55,0.15}
\usepackage{tcolorbox}
\tcbuselibrary{listings, skins, breakable}
\usepackage{enumitem}
\setlist[itemize]{topsep=0pt, leftmargin=1.5em, labelsep=0.5em}
\setlist[enumerate]{topsep=0pt, leftmargin=1.5em, labelsep=0.5em}
\usepackage{multirow}
\usepackage{graphicx}
\usepackage{colortbl}
\usepackage{subcaption}
\usepackage[export]{adjustbox} 
\usepackage{array}
\newcolumntype{C}[1]{>{\centering\arraybackslash}p{#1}}
\usepackage{tabularx}
\newcolumntype{Y}{>{\centering\arraybackslash}X}
\usepackage{marvosym}
\usepackage{xltabular} 
\usepackage{xparse}

\usepackage{fancyhdr}
\fancypagestyle{firststyle}{
    \fancyhf{}
    
    \fancyhead[L]{
        \vskip 6mm
        \raisebox{-0.27\height}{\includegraphics[width=0.12\linewidth]{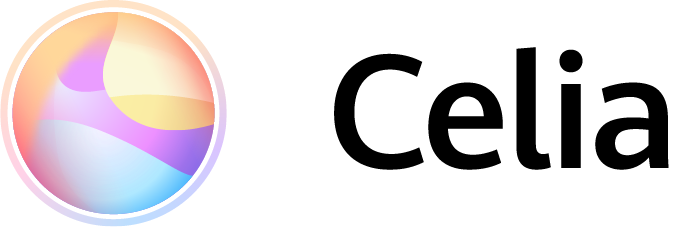}} 
        \hspace{0.4em}
        \rule[-0.43em]{1pt}{1.55em}
        \hspace{0.4em}
        Technical Report on Rubric-Based Post-Training I: Preference Optimization
    }
}

\newcommand{\zh}[1]{\begin{CJK*}{UTF8}{gbsn}#1\end{CJK*}}

\title{Visual Preference Optimization with Rubric Rewards}

\author{%
  \textbf{
  Ya-Qi Yu$^{*,\dagger}$\textsuperscript{\Letter},
  Fangyu Hong$^*$,
  Xiangyang Qu$^*$,
  Hao Wang$^*$,}\\
  \textbf{
  Gaojie Wu,
  Qiaoyu Luo,
  Nuo Xu,
  Huixin Wang,
  Wuheng Xu,}\\
  \textbf{
  Yongxin Liao,
  Zihao Chen,
  Haonan Li,
  Ziming Li,
  Dezhi Peng,}\\
  \textbf{
  Minghui Liao,
  Jihao Wu,
  Haoyu Ren,
  Dandan Tu}\\
  \footnotesize{
    $^*$~Core Contributors \;
    $^\dagger$~Project Leader \;
  }\\
  Huawei Technologies Co., Ltd.\\
}

\begin{document}

\maketitle

\thispagestyle{firststyle}

\def\thefootnote{\Letter}\footnotetext{E-mail: yuyaqi5@huawei.com}\def\thefootnote{\arabic{footnote}}

\begin{abstract}
The effectiveness of Direct Preference Optimization (DPO) depends on preference data that reflect the quality differences that matter in multimodal tasks. Existing pipelines often rely on \textit{off-policy} perturbations or \textit{coarse} outcome-based signals, which are not well suited to fine-grained visual reasoning. We propose \textit{rDPO}, a preference optimization framework based on instance-specific rubrics. For each image-instruction pair, we create a checklist-style rubric of essential and additional criteria to score responses from any possible policies. The instruction-rubric pool is built \textit{offline} and reused during the construction of \textit{on-policy} data. On public reward modeling benchmarks, rubric-based prompting massively improves a 30B-A3B judge and brings it close to GPT-5.4. On public downstream benchmarks, rubric-based filtering raises the macro average to 82.69, whereas outcome-based filtering drops it to 75.82 from 81.14. When evaluating scalability on a comprehensive benchmark, rDPO achieves 61.01, markedly outperforming the style-constrained baseline (52.36) and surpassing the 59.48 base model. Together, these results show that visual preference optimization benefits from combining on-policy data construction with instance-specific criterion-level feedback.
\end{abstract}
\section{Introduction}
\label{sec:intro}

Direct Preference Optimization~(DPO)~\cite{dpo} is now a common approach for aligning Vision-Language Models~(VLMs)~\cite{internvl-3.5,mimo-vl,keye-vl-1.5,qwen3-vl}, helping improve response quality and reduce hallucinations. Its effectiveness, however, depends critically on the quality of the underlying preference data. Early pipelines for constructing preference pairs fell into two broad categories: \textit{response-oriented} methods that inject hallucinations or exploit ``strong-weak'' model outputs~\cite{povid,ha-dpo,bpo,stic,seva,self-improve,mpo,rlaif-v}, and \textit{vision-oriented} methods that perturb visual inputs via diffusion noise or image editing~\cite{povid,mfdpo,mdpo,v-dpo,vilp,re-align,symmpo}. These methods are useful in specific settings, but they often produce \textit{off-policy} pairs. As a result, the constructed preference data can drift away from the target model's actual generation behavior.

Recent work has moved toward on-policy and self-improving pipelines. Methods such as RLAIF-V~\cite{rlaif-v} and MMPR~\cite{mpo} group model-generated responses by hallucination count or final correctness, while OPA-DPO~\cite{opa-dpo} and online DPO variants~\cite{ovip,on-policy-iterative-dpo} repeatedly sample from the latest policy during training. This line of work reduces the distribution mismatch between constructed pairs and the target model. However, most existing pipelines still rely on a reward signal of limited granularity, and they often miss differences in grounding, completeness, and reasoning quality.

To obtain richer supervision, many studies use VLM-as-a-Judge and Reward Models~(RMs) to score candidate responses~\cite{llava-rlhf,rlhf-v,wildvision,silkie,mia-dpo,mm-rlhf}. This makes large-scale data filtering practical, but it also makes judge quality the main bottleneck. Large proprietary models can provide strong annotations, yet they are expensive to use at scale. Open-source judge models~\cite{prometheus-vision,sima,llava-critic,critic-v,ixc-2.5-reward,skywork-vl-reward} are more practical, but they are often prompted with fixed templates and yield coarse scores or rankings, which lack the transparency and fine-grained feedback needed to penalize reasoning errors or hallucinations.

In the Large Language Model~(LLM) setting, rubric-based evaluation has improved both automated assessment and alignment by breaking quality into explicit criteria~\cite{llm-rubric,paperbench,carmo,rlcf,ruscarl}. In multimodal settings, however, rubric-based reward modeling remains underexplored. Concurrent work such as Omni-RRM~\cite{omni-rrm} moves in this direction, but it relies on fixed combinations of predefined criteria.

\begin{figure*}[t]
    \centering
    \includegraphics[width=\textwidth]{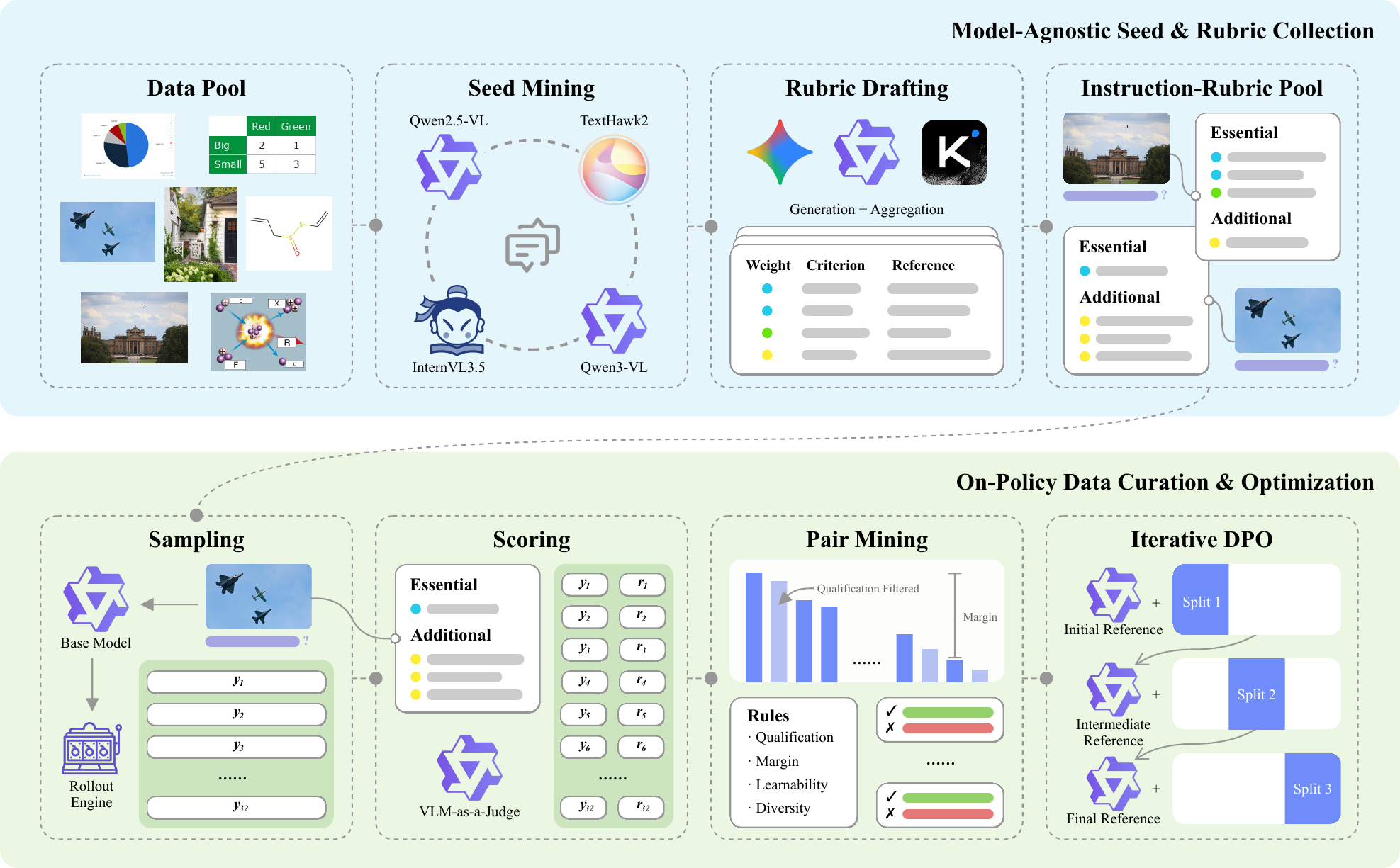}
    \caption{Overview of rDPO. The top lane constructs a model-agnostic instruction-rubric pool by mining challenging seeds via model ensembles and drafting instance-specific rubrics with structured ``essential'' and ``additional'' criteria. The bottom lane achieves on-policy data curation and optimization by sampling responses from a target policy, scoring them via rubric-grounded judging, and mining preference pairs to drive iterative DPO.}
    \label{fig:rdpo_overview}
\end{figure*}

We therefore propose rDPO, a visual preference optimization framework grounded in \textit{instance-specific} rubrics. Fig~\ref{fig:rdpo_overview} summarizes the overall pipeline. For each image-instruction pair, we build a checklist-style rubric of essential and additional criteria. These rubrics provide structured guidance that helps a moderately-sized 30B-A3B open-source judge produce more fine-grained feedback. The instruction-rubric pool is built offline and then reused during the construction of on-policy data. We evaluate the method from three angles: judge quality on public reward modeling benchmarks, a method-validation setting where rubric-based filtering outperforms outcome-based filtering under the same on-policy setup, and a scaling-validation setting on a comprehensive in-house benchmark.

In summary, the main contributions of this work are as follows:
\begin{itemize}
    \item We propose rDPO, a visual preference optimization framework based on instance-specific rubric-based reward modeling. The method differs from template-based rubric composition by generating per-instance criteria for visual preference construction.
    \item On public reward-modeling benchmarks, instance-specific rubrics improve the judgment quality of a moderately-sized 30B-A3B open-source judge without additional training.
    \item We construct a 300K instruction-rubric pool and provide an automated pipeline for generating on-policy preference data for target-policy-specific training.
    \item In method validation, outcome-based filtering reduces the macro average from 81.14 to 75.82, while rDPO raises it to 82.69. In scaling validation, rDPO achieves 61.01, markedly outperforming the style-constrained baseline (52.36) and surpassing the 59.48 base model.
\end{itemize}
\section{Related Works}
\label{sec:related}

\subsection{Reward Models for Multimodal Alignment}
Aligning VLMs with complex human intent has motivated the curation of large-scale multimodal preference datasets. Early datasets relied on extensive human feedback, including LLaVA-RLHF~\cite{llava-rlhf}, RLHF-V~\cite{rlhf-v}, MM-RLHF~\cite{mm-rlhf}, WildVision~\cite{wildvision}, and VisionArena~\cite{visionarena}. To reduce annotation costs, AI feedback mechanisms followed, as seen in VLFeedback~\cite{silkie}, RLAIF-V~\cite{rlaif-v}, MIA-DPO~\cite{mia-dpo}, and LLaVA-Critic~\cite{llava-critic}. In parallel, specialized visual RMs have been developed as automated evaluators. Early explorations like Prometheus-Vision~\cite{prometheus-vision}, SIMA~\cite{sima}, and LLaVA-Critic~\cite{llava-critic} established the foundation. Recent work has further refined RM capabilities: CAREVL~\cite{carevl} distills language reward knowledge into visual RMs, SVIP-Reward~\cite{svip-reward} incorporates stepwise visual programs, and comprehensive models like MM-RLHF-Reward~\cite{mm-rlhf}, IXC-2.5-Reward~\cite{ixc-2.5-reward}, and Skywork-VL Reward~\cite{skywork-vl-reward} integrate diverse multimodal preference corpora. Despite this progress, existing RMs typically produce scalar scores or rely on rigid prompting templates, and lack the transparency and multi-dimensional granularity needed to guide nuanced visual reasoning.

\subsection{Rubric-Grounded Evaluation and Alignment}
Moving from generic scoring to rubric-based frameworks has delivered substantial improvements in LLM evaluation and alignment. For evaluation, customized rubrics have proven essential for complex, multi-dimensional tasks~\cite{llm-rubric,paperbench,healthbench,profbench,prbench}. In post-training, frameworks such as Rubrics-as-Rewards~\cite{rlrr}, Rubicon~\cite{rubicon}, and CARMO~\cite{carmo} use rubrics to provide structured reward signals. To generate discriminative rubrics, methods like RLCF~\cite{rlcf}, OpenRubrics~\cite{openrubrics}, and Rubrichub~\cite{rubric-hub} employ contrastive generation. Dynamic refinement processes are introduced by OnlineRubrics~\cite{online-rubrics}, Auto-Rubric~\cite{auto-rubrics}, and Rubric-ARM~\cite{rubric-arm} to mitigate reward over-optimization. RuscaRL~\cite{ruscarl} further introduces rubrics for both reward modeling and hint-guided rollout generation.

In the multimodal domain, rubric-based alignment remains in its early stages. Recent evaluation benchmarks like JudgeAnything~\cite{judge-anything} and Multi-Crit~\cite{multi-crit} have introduced instance-specific checklists for multi-criteria assessment, highlighting the demand for structured evaluation. For model alignment, the concurrent work Omni-RRM~\cite{omni-rrm} represents an early attempt to provide fine-grained feedback. However, it relies on fixed combinations of predefined rubric templates. Our method differs in that it generates instance-specific criteria for preference pair mining during data construction.
\section{Preliminaries}
\label{sec:prem}

\subsection{DPO}
\label{ssec:dpo}

This work centers on self-improvement through preference alignment, adopting DPO~\cite{dpo} as the primary framework. DPO obviates explicit reward modeling by leveraging the analytical mapping between the reward function and the optimal policy under the Bradley-Terry preference model~\cite{bradley-terry}.

Consider a dataset $\mathcal{D}$ consisting of preference triplets $(x, y_c, y_r)$, where $x$ denotes the input (comprising both image and text prompts), while $y_c$ and $y_r$ represent the chosen and rejected responses, respectively. The DPO objective is defined as:
\begin{equation}\label{eq:dpo}
    \mathcal{L}_\text{DPO}(\pi_{\theta}; \pi_0) = -\log \sigma \left(\beta \log \frac{\pi_{\theta}(y_c \mid x)}{\pi_0(y_c \mid x)} - \beta \log \frac{\pi_{\theta}(y_r \mid x)}{\pi_0(y_r \mid x)}\right),
\end{equation}
where $\sigma$ is the sigmoid function, $\beta$ is the KL penalty coefficient, and $\pi_\theta$ represents the policy model parameterized by $\theta$, which is initialized from the reference model $\pi_0$.

The derivation~\cite{dpo} of the gradient of the DPO loss with respect to $\theta$ can be written as follows:
\begin{equation}\label{eq:dpo_gradient}
    \nabla_\theta \mathcal{L}_\text{DPO}(\pi_\theta; \pi_0) = - \beta \underbrace{\sigma(\hat{r}_\theta(x, y_r) - \hat{r}_\theta (x, y_c))}_\text{scaling by reward margin} \bigg[\underbrace{\nabla_\theta \log \pi(y_c \mid x)}_\text{increase likelihood of $y_c$} - \underbrace{\nabla_\theta \log \pi(y_r \mid x)}_\text{decrease likelihood of $y_r$}\bigg],
\end{equation}
where $\hat{r}_\theta(x, y) = \beta \log \frac{\pi_\theta(y \mid x)}{\pi_0(y \mid x)}$ represents the implicit reward. This gradient structure intuitively scales the updates based on the reward margin, increasing the likelihood of $y_c$ while penalizing $y_r$.

\subsection{MPO}
\label{ssec:mpo}

To integrate preference learning with the stability of Supervised Fine-Tuning~(SFT), we introduce a simplified version of Mixed Preference Optimization~(MPO)~\cite{mpo}. This formulation augments the DPO objective with an auxiliary SFT loss:
\begin{equation}\label{eq:mpo}
    \mathcal{L}_\text{MPO}(\pi_{\theta}; \pi_0) = \mathcal{L}_\text{DPO}(\pi_{\theta}; \pi_0) + \mathcal{L}_\text{SFT}(\pi_{\theta}).
\end{equation}

Specifically, the SFT term reinforces the chosen response $y_c$, acting as a form of rejection sampling:
\begin{equation}\label{eq:sft}
    \mathcal{L}_\text{SFT}(\pi_{\theta}) = -\alpha \log \sigma\left(\log \pi_{\theta}(y_c \mid x)\right),
\end{equation}
where $\alpha$ is a scaling coefficient. Here, we \textit{exclude the length normalization} to ensure the magnitude remains consistent with the DPO term.

The resulting gradient:
\begin{equation}\label{eq:sft_gradient}
    \nabla_\theta \mathcal{L}_\text{SFT}(\pi_\theta) = - \alpha \underbrace{\nabla_\theta\log \pi(y_c \mid x)}_\text{increase likelihood of $y_c$},
\end{equation}
effectively increases the weight of the gradient on the chosen response to prevent policy collapse.

\subsection{Rubric}
\label{ssec:rubric}

To enhance the self-improvement cycle, we sample on-policy responses $y \sim \pi_0(\cdot|x)$ and introduce a rubric-based RM for evaluation. Scoring against the target model's own outputs reduces the mismatch introduced by static offline datasets.

Specifically, our framework employs Generative RMs~(GenRMs) that evaluate responses against a structured, instance-specific checklist. For each input $x \in \mathcal{X}$, we define a localized rubric encompassing $K$ distinct criteria (e.g., factual accuracy, instruction following), denoted as a set $C^x$:
\begin{equation}\label{eq:rubric_vector}
    C^x = \left\{c^x_1, c^x_2, \dots, c^x_K\right\}.
\end{equation}

Formally, the reward model $r_\phi$ is defined as a function that takes an input $x \in \mathcal{X}$, a generated response $y \in \mathcal{Y}$, and the corresponding rubric $C^x \in \mathcal{C}$ to yield a multi-dimensional feedback vector $\mathbf{s} \in \mathbb{R}^K$. Compared to traditional scalar rewards, this formulation provides greater transparency and interpretability. Furthermore, it enables fine-grained data filtering and facilitates the construction of nuanced preference pairs for subsequent optimization.
\section{Methodology}
\label{sec:method}

Our data construction pipeline consists of three stages: (1) \textit{Seed Mining}, which filters high-quality seed instructions from a broader data pool; (2) \textit{Rubric Drafting}, which generates instance-specific rubrics for the selected seeds; and (3) \textit{Preference Data Curation}, which samples on-policy responses and constructs preference pairs for a specific target policy model. The first two stages are \textit{model-agnostic}: the resulting instruction-rubric pool can be reused across different target models.

\subsection{Seed Mining}
\label{ssec:prep_seed}

To curate a high-quality and challenging set of instructions, we employ a disagreement-based filtering mechanism, which operates in two steps:

\textbf{Large-Scale Rollout.} We first conduct a large-scale parallel rollout to collect initial responses. To capture diverse model behaviors, we use an ensemble of moderately-sized models spanning both dense and Mixture-of-Experts (MoE) architectures: TextHawk2-7B~\cite{texthawk2}, Qwen2.5-VL-7B~\cite{qwen2.5-vl}, InternVL3.5-30B-A3B~\cite{internvl-3.5}, and Qwen3-VL-30B-A3B~\cite{qwen3-vl}.

\textbf{Disagreement-based Filtering.} We then apply a reference-free, disagreement-based filtering strategy. We retain only those instances where the ensemble models fail to reach a consensus. Such disagreement often identifies complex, ambiguous, or edge cases that are highly informative for preference optimization. Finally, to maintain domain balance, we apply uniform sampling across all original data sources to construct the final seed set.

\subsection{Rubric Drafting}
\label{ssec:prep_rubric}

To ensure that the evaluation criteria are mutually exclusive and strictly verifiable by an RM, our construction process adheres to four core principles: (1) \textit{Atomic:} Each criterion targets a single, indivisible key point or sub-query within the original instruction. (2) \textit{Comprehensive:} The criteria list jointly covers all vital dimensions of the user query, so that no critical aspect of a complete response is overlooked. (3) \textit{Precise:} The evaluation aligns strictly with the user query, avoiding redundant checks or extraneous information. (4) \textit{Objective:} Assessments are grounded in observable facts, empirical evidence, or reference answers, eliminating variance from subjective interpretations.

Based on these principles, we explicitly categorize the criteria into two distinct types: \textit{essential} and \textit{additional}. \textbf{Essential} criteria capture the core information prioritized by the query; satisfying these is a prerequisite for a conceptually sound response. \textbf{Additional} criteria encompass relevant image facts, supplementary knowledge, or intermediate steps required to derive the answer.

Formally, each check item is structured as a triplet comprising a \textit{criterion}, a \textit{reference}, and a fixed-point \textit{weight}. The \textbf{criterion} dictates a concrete and observable assertion to be verified. The \textbf{reference} serves as the ground truth, strictly derived from image facts. Finally, the \textbf{weight} quantifies the item's importance across three discrete levels: \textit{1 (Auxiliary)} for helpful but non-critical information; \textit{2 (Important)} for content that significantly enhances the user experience; and \textit{3 (Key)} for critical elements where any omission or deviation constitutes a definitive error.

To generate rubrics following this schema, we employ frontier reasoning models to generate instance-specific checklists in JSON format (see Appendix~\ref{sec:prompt}). To account for varying annotation quality across seed datasets, we adapt our generation strategy accordingly:

\textbf{Expert-Grounded Generation.} For datasets with trustworthy human annotations, the rubric generation is explicitly conditioned on the provided ground truths. This ensures that the resulting rubrics are firmly anchored in expert knowledge and factual accuracy.

\textbf{Answer-Agnostic Generation.} Conversely, for datasets characterized by noisy or missing annotations, we deliberately exclude the original answers from the prompt. This prevents the generation process from being misled by subpar reference responses.

Furthermore, relying on a single model for generation risks introducing systematic bias or persistent perception errors. To mitigate this and expand coverage, we independently prompt multiple models and synthesize their candidate rubrics into a unified checklist via a secondary aggregation prompt.

\subsection{Preference Data Curation}
\label{ssec:prep_pair}

In this stage, we sample and score multiple responses per instruction from the target policy using the instruction-rubric pool. This rollout yields a diverse candidate set for preference pair construction.

\subsubsection{Reward Modeling}
\label{sssec:rubric_rm}

\textbf{Scoring.} To evaluate the generated candidates, we employ rubric-grounded VLM-as-a-Judge via zero-shot prompting. For each criterion specified in the rubric, the judge assigns a discrete score $s \in \{0, 0.5, 1\}$, corresponding to \textit{no credit}, \textit{partial credit}, and \textit{full credit}, respectively. To facilitate reliable parsing, the judge is instructed to output the evaluation in JSON format (see Appendix~\ref{sec:prompt}).

\textbf{Voting.} To reduce the variance and hallucination of VLM-as-a-Judge, we run the scoring process three times independently for each response and adopt the median score for each criterion.

\textbf{Aggregation.} The criterion-level scores are aggregated to compute the overall reward for a given input $x$ and response $y$. Assume each criterion $c^x_k$ is associated with a reference answer $a^x_k$ and a weight $w^x_k$, we compute the overall reward as follows:
\begin{equation}
    \label{eq:reward_agg}
    r_\phi(x, y, C^x) = \sum_k w^x_k \cdot s_\phi(x,y,c^x_k,a^x_k).
\end{equation}

\subsubsection{Pair Mining}
\label{sssec:filter_pair}

Once all candidate responses are scored, we construct the final preference pairs by applying the following filtering criteria. This guarantees that the reward margin reflects a genuine and interpretable difference in quality. The pairing process strictly adheres to the following rules:

\begin{itemize}
    \item \textbf{Rule 1 (Chosen Qualification):} The chosen response $y_c$ must meet basic correctness thresholds. It must receive full credit ($s=1$) on most essential criteria, allowing at most one partial credit ($s=0.5$). Furthermore, responses exhibiting repetition loops or language mixing are discarded.
    \item \textbf{Rule 2 (Rejected Qualification):} The rejected response $y_r$ must exhibit definitive flaws. It must either fail ($s=0$) at least one essential criterion or receive partial credit ($s=0.5$) on at least two.
    \item \textbf{Rule 3 (Margin Constraint):} To ensure a meaningful qualitative gap, the overall reward difference must satisfy a margin constraint: $r_\phi(x, y_c, C^x) - r_\phi(x, y_r, C^x) \geq \delta$, where $\delta > 0$ is an empirical margin hyperparameter to ensure a distinct quality gap.
    \item \textbf{Rule 4 (Maximum Learning Capacity):} Among all valid pairs satisfying the above criteria for a given instruction, we select the top four pairs $(y_c, y_r)$ that maximize the overall reward margin.
    \item \textbf{Rule 5 (Diversity Control):} To prevent the policy model from overfitting, each unique response is allowed to appear a maximum of twice across all preference pairs in the final dataset.
\end{itemize}
\section{Experiment}
\label{sec:expr}

\subsection{Evaluation Results of Reward Modeling}
\label{ssec:expr_rm}

\textbf{Benchmarks.} To validate the effectiveness of the proposed rubrics in multimodal reward modeling, we compare it against established visual judges across several preference benchmarks. These encompass Multimodal RewardBench~(MM-RB)~\cite{mm-rewardbench}, VL-RewardBench~(VL-RB)~\cite{vl-rewardbench}, MLLM-as-a-Judge~(MaaJ)~\cite{mllm-as-a-judge}, VisionArena-Battle~(VA-B)~\cite{visionarena}, and WildVision-Battle~(WV-B)~\cite{wildvision}. Across all benchmarks, we exclude tie samples to focus on clear preferences. Furthermore, we sub-sample 1,200 pairs from VA-B and WV-B, respectively, to optimize evaluation efficiency.

\textbf{Baselines.} To ensure a rigorous evaluation, we benchmark our approach against a wide range of state-of-the-art models: (1) Proprietary Frontier VLMs, such as Claude 4.6 Opus~\cite{claude-opus-4.6}, GPT-5.4~\cite{gpt-5.4}, and the Gemini series~\cite{gemini3}; (2) Open-source Generalist VLMs, specifically the Qwen3-VL series~\cite{qwen3-vl}; and (3) Open-source Specialist RMs, including IXC-2.5-Reward~\cite{ixc-2.5-reward} and Skywork-VL Reward~\cite{skywork-vl-reward}.

\begin{table*}[t]
    \centering
    \caption{Comprehensive evaluation results across diverse preference benchmarks. Accuracy (\%) is reported. \textit{+ Rubric} denotes our proposed strategy. \textit{Macro} presents a triplet: (Full Average / Average excluding MM-RB \& VL-RB / Average excluding VA-B \& WV-B). \textcolor{gray!70}{Gray} numbers indicate potential data contamination. If a model has contaminated entries, the corresponding macro averages are ignored (-). The best performance in each section is \textbf{bolded}, and the second best is \underline{underlined}.}
    \label{tab:rm_bench}
    \resizebox{\textwidth}{!}{
    \begin{tabular}{l ccccc c}
        \toprule
        \textbf{Model Evaluator} & \textbf{MM-RB} & \textbf{VL-RB} & \textbf{MaaJ} & \textbf{VA-B} & \textbf{WV-B} & \textbf{Macro (Full / - RB / - B)} \\
        \midrule

        \multicolumn{7}{l}{\textit{\textbf{Proprietary Frontier VLMs}}} \\
        \midrule
        Claude 4.6 Opus & \textbf{82.23} & \underline{74.42} & \textbf{71.39} & \textbf{78.83} & \textbf{74.00} & \textbf{76.17} / \textbf{74.74} / \underline{76.01} \\
        GPT-5.4 & \underline{81.91} & \textbf{78.27} & \underline{70.35} & 74.25 & 69.75 & \underline{74.91} / 71.45 / \textbf{76.84} \\
        Gemini 3.1 Pro & \textcolor{gray!50}{88.79} & \textcolor{gray!50}{86.45} & 66.56 & 76.25 & \underline{73.33} & \quad - \quad / 72.05 / \quad - \ \ \,\ \\
        Gemini 3.0 Flash & \textcolor{gray!50}{87.90} & \textcolor{gray!50}{85.89} & 66.60 & \underline{76.66} & 73.42 & \quad - \quad / \underline{72.23} / \quad - \ \ \,\ \\
        Gemini 3.1 Flash-Lite & \textcolor{gray!50}{82.25} & \textcolor{gray!50}{81.40} & 64.31 & 74.33 & 67.00 & \quad - \quad / 68.55 / \quad - \ \ \,\ \\
        \midrule

        \multicolumn{7}{l}{\textit{\textbf{Open-source Generalist VLMs}}} \\
        \midrule
        Qwen3-VL-30B-A3B-Instruct & 75.82 & 65.52 & 70.74 & 74.08 & 72.50 & 71.73 / 72.44 / 70.69 \\
        \quad \textit{+ CoT} & 72.49 & 60.22 & \underline{70.20} & 72.17 & 70.83 & 69.18 / 71.07 / 67.64 \\
        \rowcolor{blue!5}
        \quad \textit{+ Rubric} & \underline{82.06} & \textbf{73.54} & 69.13 & \underline{75.50} & \underline{73.92} & \underline{74.83} / \underline{72.85} / \underline{74.91} \\
        \addlinespace
        Qwen3-VL-32B-Instruct & 79.35 & 70.73 & \textbf{72.18} & 75.42 & 71.83 & 73.90 / 73.14 / 74.09 \\
        \quad \textit{+ CoT} & 81.19 & 72.15 & 71.83 & 74.58 & 72.50 & 74.45 / 72.97 / \textbf{75.06} \\
        \rowcolor{blue!5}
        \quad \textit{+ Rubric} & \textbf{83.02} & \underline{72.49} & 68.93 & \textbf{76.83} & \textbf{74.00} & \textbf{75.05} / \textbf{73.25} / 74.81 \\
        \midrule

        \multicolumn{7}{l}{\textit{\textbf{Open-source Specialist RMs}}} \\
        \midrule
        IXC-2.5-Reward & 69.12 & 66.64 & 69.15 & \textcolor{gray!50}{86.25} & \textcolor{gray!50}{89.83} & \quad - \quad / \quad - \quad / 68.30 \\
        Skywork-VL Reward & 74.25 & 73.54 & 59.93 & 71.67 & 65.08 & 68.89 / 65.56 / 69.24 \\
        \bottomrule
    \end{tabular}
    }
\end{table*}

Table~\ref{tab:rm_bench} shows that rubric prompting improves both Qwen3-VL judges over their vanilla prompts, whereas CoT prompting does not consistently help. For Qwen3-VL-30B-A3B, rubric prompting raises the overall macro average from 71.73 to 74.83. For Qwen3-VL-32B, it raises the macro average from 73.90 to 75.05. The gray entries indicate likely contamination in some baselines, so we do not compare full macro averages for those settings. These results validate that instance-specific rubrics consistently enhance zero-shot judge quality, consistently lifting Qwen3-VL judges and bringing the 30B-A3B model close to GPT-5.4 on the reported suite (74.83 vs. 74.91).

\subsection{Evaluation Results of Preference Optimization}
\label{ssec:expr_po}

\textbf{Data Curation.} We construct two downstream preference datasets and study them in two complementary settings. The first is a method-validation setting designed to isolate the effect of rubric-based data curation. Here we use the training splits of AI2D~\cite{ai2d}, ChartQA~\cite{chartqa}, and M3CoT~\cite{m3cot} as seed sources, evaluate the resulting policies on the corresponding test sets, and compare outcome-based filtering against rubric-based filtering under the same base policy and training recipe. The second is a scaling-validation setting designed to test the pipeline in a broader multi-task regime. Here we use the seed pool described in Section~\ref{ssec:prep_seed} and evaluate the resulting policy on our comprehensive in-house benchmark under concise-response constraints. In both settings, we sample 32 candidate responses per prompt and score and filter them with the procedure in Section~\ref{sssec:filter_pair} to construct model-specific preference pairs. Prompts that yield no valid pairs are discarded.

\textbf{Training Details.} We select Qwen3-VL-30B-A3B-Instruct as the initial policy. The optimization is driven by AdamW with a constant learning rate of $1 \times 10^{-6}$, $\beta = 0.5$, and $\alpha = 0$. We use a global batch size of 128 for the method-validation setting and 512 for the scaling-validation setting. For the latter, we further divide the constructed dataset into three splits and apply iterative DPO~\cite{on-policy-iterative-dpo}.

\begin{table}[t]
    \centering
    \caption{Results in the method-validation setting on AI2D, ChartQA, and M3CoT. Note that for the ChartQA evaluation, we adopt LLM-as-a-Judge, which is stricter than relaxed accuracy. For the latter two rows, each entry reports the metric followed by its absolute change relative to the base model.}
    \label{tab:pub_vlm_bench}
    \resizebox{\textwidth}{!}{
    \begin{tabular}{lllll}
        \toprule
        \textbf{Policy Model} & \textbf{AI2D} & \textbf{ChartQA} & \textbf{M3CoT} & \textbf{Macro Avg.} \\
        \midrule
        Qwen3-VL-30B-A3B-Instruct & 84.10 & 82.32 & 77.00 & 81.14 \\
        \quad \textit{+ Outcome-based Filtering} & 78.53 \, {\footnotesize\textcolor{red!70!black}{$-5.57$}} & 75.48 \, {\footnotesize\textcolor{red!70!black}{$-6.84$}} & 73.45 \, {\footnotesize\textcolor{red!70!black}{$-3.55$}} & 75.82 \, {\footnotesize\textcolor{red!70!black}{$-5.32$}} \\
        \rowcolor{blue!5}
        \quad \textit{+ Rubric-based Filtering (Ours)} & 85.95 \, {\footnotesize\textcolor{green!60!black}{$+1.85$}} & 83.02 \, {\footnotesize\textcolor{green!60!black}{$+0.70$}} & 79.11 \, {\footnotesize\textcolor{green!60!black}{$+2.11$}} & 82.69 \, {\footnotesize\textcolor{green!60!black}{$+1.55$}} \\
        \bottomrule
    \end{tabular}
    }
\end{table}

Table~\ref{tab:pub_vlm_bench} summarizes the method-validation setting. Outcome-based filtering lowers performance on all three tasks relative to the base model: AI2D drops from 84.10 to 78.53, ChartQA from 82.32 to 75.48, and M3CoT from 77.00 to 73.45, reducing the macro average from 81.14 to 75.82. In contrast, rubric-based filtering improves all three metrics over the base model, reaching 85.95 on AI2D, 83.02 on ChartQA, and 79.11 on M3CoT, with a macro average of 82.69. These results suggest that on-policy data alone is not sufficient in this setting, and the gains depend on pairing those rollouts with instance-specific criterion-level feedback.

\begin{table}[t]
    \centering
    \caption{Results in the scaling-validation setting on the comprehensive benchmark before and after DPO. The initial policy is Qwen3-VL-30B-A3B-Instruct. \textit{+ Prompting} denotes the baseline explicitly instructed to generate concise responses, mitigating the default model's verbosity and markdown abuse. \textit{+ rDPO} denotes the model aligned using our generated preference dataset under the concise constraint. Absolute performance gains ($\Delta$) represent improvements over the \textit{Prompting} baseline.}
    \label{tab:vlm_bench}
    \resizebox{\textwidth}{!}{
    \begin{tabular}{l cccc c}
        \toprule
        \multirow{2}{*}{\textbf{Policy Model}} & \multicolumn{2}{c}{\textbf{Basic Capabilities}} & \multicolumn{2}{c}{\textbf{Advanced Cognition}} & \multirow{2}{*}{\textbf{Avg.}} \\
        \cmidrule(lr){2-3} \cmidrule(lr){4-5}
        & \textbf{Perception} & \textbf{Understanding} & \textbf{Knowledge} & \textbf{Reasoning} & \\
        \midrule
        Qwen3-VL-30B-A3B-Instruct & 56.26 & 74.10 & \textbf{71.94} & 51.53 & 59.48 \\
        \quad \textit{+ Prompting} & 55.67 & 68.57 & 61.02 & 42.86 & 52.36 \\
        \rowcolor{blue!5}
        \quad \quad \textit{+ rDPO (Ours)} & \textbf{62.82} & \textbf{77.00} & 71.27 & \textbf{51.62} & \textbf{61.01} \\
        \midrule
        \textit{Absolute Gain ($\Delta$)} & \textcolor{green!60!black}{\textit{+7.15}} & \textcolor{green!60!black}{\textit{+8.43}} & \textcolor{green!60!black}{\textit{+10.25}} & \textcolor{green!60!black}{\textit{+8.76}} & \textcolor{green!60!black}{\textit{+8.65}} \\
        \bottomrule
    \end{tabular}
    }
\end{table}

As shown in Table~\ref{tab:vlm_bench}, the default baseline achieves an average score of 59.48, though it tends to produce overly verbose responses with excessive Markdown formatting. Attempting to control this verbosity via concise system prompts causes the score to drop to 52.36. In contrast, by training on conciseness-constrained on-policy data, rDPO reaches a superior score of 61.01 and delivers an 8.65 point improvement over the prompting baseline.

Compared with the original base model, rDPO is slightly higher on average (61.01 vs. 59.48) and improves perception and understanding, while the knowledge dimension remains slightly lower than the base model (71.27 vs. 71.94). Together with the method-validation results, this supports the view that on-policy data becomes more effective when paired with rubric-guided criterion-level scoring. In the scaling-validation setting, rDPO recovers the drop introduced by prompt-only conciseness enforcement and slightly improves average score over the base model.

\subsection{Ablation Study}
\label{ssec:ablation}

\begin{figure*}[t]
    \centering
    \begin{subfigure}[b]{0.248\textwidth}
        \centering
        \includegraphics[width=\linewidth]{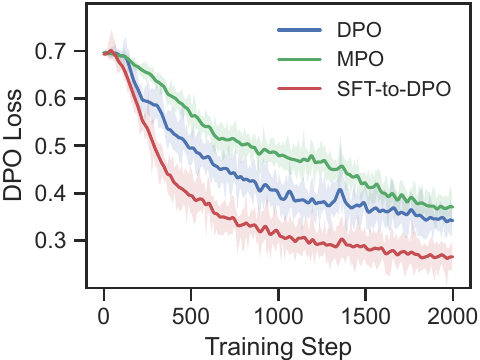}
        \caption{DPO Loss}
        \label{fig:loss}
    \end{subfigure}%
    \begin{subfigure}[b]{0.248\textwidth}
        \centering
        \includegraphics[width=\linewidth]{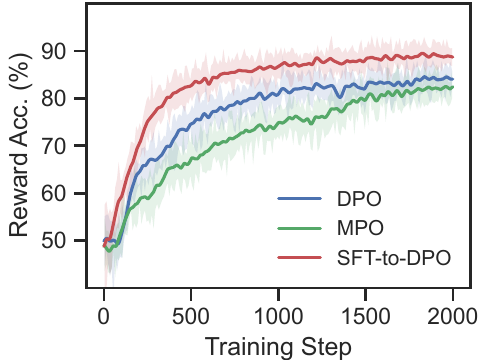}
        \caption{Reward Accuracy}
        \label{fig:acc}
    \end{subfigure}%
    \begin{subfigure}[b]{0.248\textwidth}
        \centering
        \includegraphics[width=\linewidth]{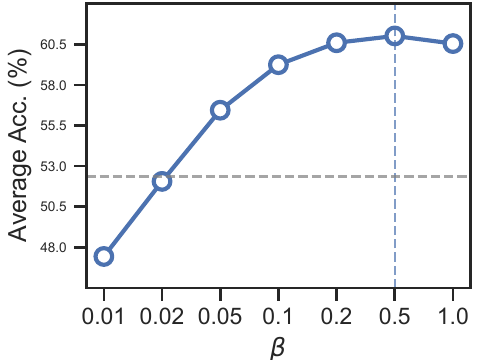}
        \caption{KL Penalty}
        \label{fig:beta}
    \end{subfigure}%
    \begin{subfigure}[b]{0.248\textwidth}
        \centering
        \includegraphics[width=\linewidth]{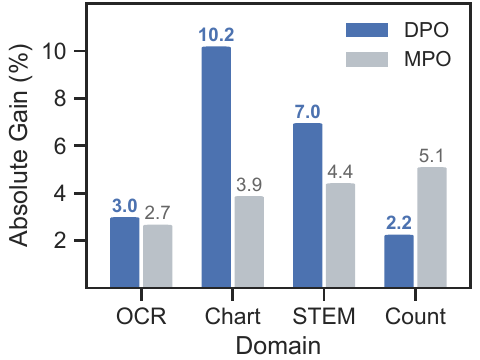}
        \caption{Domain Performance}
        \label{fig:alpha}
    \end{subfigure}
    \caption{Training dynamics and hyperparameter ablations. (a)(b) Convergence curves for three training paradigms on off-policy data, showing loss and reward accuracy. (c) Effect of varying the KL penalty coefficient $\beta$. (d) Effect of the SFT loss broken down by task domain.}
\end{figure*}

\textbf{Ablation Study of Policy Alignment.}
To investigate the necessity of on-policy data during alignment, we analyze preference optimization under severe distribution shifts using the MMPR dataset~\cite{mpo}, which was originally gathered for the InternVL series~\cite{internvl-3.5}. We evaluate three different training configurations on the Qwen3-VL-30B-A3B-Instruct model:
\begin{itemize}
    \item \textbf{DPO (Off-Policy):} Standard DPO on MMPR with the base checkpoint as $\pi_0$ and KL penalty coefficient $\beta = 0.001$, leaving the full distributional mismatch unresolved.
    \item \textbf{MPO (Off-Policy):} MPO training with an identical SFT scaling coefficient $\alpha = 0.001$; the SFT term provides additional supervision on chosen responses but does not alter $\pi_0$.
    \item \textbf{SFT-to-DPO (On-Policy Approximation):} A two-stage progressive approach designed to simulate on-policy optimization. The model first undergoes SFT on the chosen responses; the resulting checkpoint $\pi_{\mathrm{sft}}$ then serves as the reference model for subsequent DPO training, thereby narrowing the distribution gap between the policy and the preference data.
\end{itemize}

Fig.~\ref{fig:loss} and Fig.~\ref{fig:acc} show clear differences in convergence. Standard DPO converges slowly because the reference policy and off-policy data distribution are mismatched, leading to unstable gradients. MPO’s loss drops unevenly—starting fast, then slowing down, before speeding up again—which suggests the policy is shifting throughout the process. In contrast, SFT-to-DPO is the most stable, achieving the lowest final loss and the highest reward accuracy.

Our analysis reveals that DPO is inherently sensitive to distribution shift, which degrades the accuracy of the implicit reward. The ``inflection point'' observed during MPO training likely indicates an initial phase where the model first resolve formatting discrepancies before it can effectively learn from preference signals. The SFT-to-DPO pipeline avoids this bottleneck by ensuring the reference policy matches the data distribution, leading to more stable gradient updates throughout training.

\textbf{Ablation on the KL Penalty Coefficient.}
Contrary to previous literature, we empirically find that the KL penalty coefficient reaches an optimal trade-off at $\beta = 0.5$ (Fig.~\ref{fig:beta}). Maintaining a substantial penalty anchors the policy to the reference distribution, mitigating policy collapse by preventing the reduction of chosen-response likelihoods~\cite{smaug} or implicit reward overfitting~\cite{length-regularized-dpo}.

\textbf{Ablation on the SFT Scaling Coefficient.}
Figure~\ref{fig:alpha} compares the greatest performance gain of DPO and MPO across different domains. We observe that the effect of the SFT loss depends on the specific task. In most cases, MPO leads to worse results. Although SFT loss improves training stability, it may cause overfitting on low-complexity samples. However, MPO remains effective for tasks where the base model has undesirable priors. A typical example is counting, where giving the final answer before counting step-by-step reduces accuracy. In these cases, the SFT loss acts as a constraint that aligns the model with the high-quality distribution obtained through rejection sampling.

\begin{table}[t]
    \centering
    \caption{Comprehensive ablation studies on our in-house benchmark. The first row represents our full \textit{rDPO} pipeline (the baseline for this ablation). Subsequent rows denote variants where specific alignment modules are either removed or altered.}
    \label{tab:ablation}
    \resizebox{\textwidth}{!}{
    \begin{tabular}{l cccc c}
        \toprule
        \multirow{2}{*}{\textbf{Ablation Variant}} & \multicolumn{2}{c}{\textbf{Basic Capabilities}} & \multicolumn{2}{c}{\textbf{Advanced Cognition}} & \multirow{2}{*}{\textbf{Avg.}} \\
        \cmidrule(lr){2-3} \cmidrule(lr){4-5}
        & \textbf{Perception} & \textbf{Understanding} & \textbf{Knowledge} & \textbf{Reasoning} & \\
        \midrule
        \rowcolor{blue!5}
        \textit{rDPO (Full Pipeline)} & 62.82 & 77.00 & 71.27 & 51.62 & \textbf{61.01} \\
        \midrule
        \quad Single-stage DPO & 63.02 & 75.73 & 71.27 & 48.62 & 59.27 \\
        \midrule
        \quad w/o essential qualification & 60.76 & 75.20 & 67.93 & 48.62 & 58.36 \\
        \midrule
        \quad Random reward margin ($\geq \delta$) & 63.22 & 77.69 & 71.27 & 50.74 & 60.76 \\
        \quad Minimum reward margin ($\geq \delta$) & 62.03 & 76.22 & 69.71 & 50.52 & 59.96 \\
        \midrule
        \quad w/o diversity control & 61.95 & 76.73 & 70.14 & 49.00 & 59.34 \\
        \bottomrule
    \end{tabular}
    }
\end{table}

\textbf{Ablation on Iterative Optimization.}
We compare the standard single-stage DPO with an iterative paradigm (i.e., the full rDPO pipeline), in which the preference dataset is partitioned into three sequential training splits. As shown in Table~\ref{tab:ablation}, iterative DPO achieves better overall performance (61.01 vs. 59.27). In this setup, progressively updating the reference model on partitioned data helps calibrate the implicit reward, thereby leading to enhanced model performance.

\textbf{Ablation on Essential Qualification.}
We compare our approach against a baseline lacking the essential/additional criteria distinction (Rules 1 \& 2). As shown in Table~\ref{tab:ablation}, enforcing this qualification rule improves average performance from 58.36 to 61.01. This pattern suggests that separating fundamental errors from additional criteria helps produce cleaner preference pairs in our setup.

\textbf{Ablation on Learning Capacity.}
We compare models trained on datasets constructed using maximum, minimum, and random reward margins (Rule 3 \& 4, all strictly bounded by $\delta = 5$ to prevent tie samples). Table~\ref{tab:ablation} shows that selecting pairs with the maximum reward margin gives the best overall performance (61.01 vs. 59.96 vs. 60.76). In this setting, larger reward margins appear to provide a more useful training signal than smaller or randomly selected margins.

\textbf{Ablation on Diversity Control.}
We evaluate the impact of limiting response frequency (Rule 5). Removing this constraint degrades the average performance to 59.34, suggesting that reducing exposure to repetitive response patterns enhances downstream performance.
\section{Conclusion and Limitations}
\label{sec:conclude}

We presented rDPO, a visual preference optimization framework built on instance-specific rubric-based reward modeling. The pipeline separates offline instruction-rubric pool construction from rollout, scoring, and on-policy data curation.

On public reward modeling benchmarks, rubric prompting improves both Qwen3-VL judges over their vanilla prompts, whereas CoT prompting does not consistently help. It brings the moderately-sized 30B-A3B judge close to GPT-5.4 on the reported suite (74.83 vs. 74.91). In the method-validation setting, under the same on-policy rollout setup, outcome-based filtering lowers the macro average from 81.14 to 75.82, while rubric-based filtering raises it to 82.69. When evaluating scalability on our comprehensive benchmark, rDPO achieves 61.01, markedly outperforming the style-constrained baseline (52.36) and surpassing the 59.48 base model. Taken together, these results suggest that effective visual preference optimization requires not only on-policy data, but also instance-specific criterion-level feedback.

\paragraph{Limitations and future work.} The current pipeline still consumes rubric feedback indirectly: instance-specific criteria are used to construct preference pairs, and the policy is then optimized with iterative DPO. This design is stable and modular, but it leaves the criterion-level signal outside the optimization loop. A natural next step is to move toward more online training, where rubric-guided rewards can evolve with the policy itself. In particular, extending rubric-based supervision to Group Relative Policy Optimization~(GRPO)~\cite{grpo} may offer a more direct way to optimize groups of sampled responses against structured criteria. We leave this tighter integration of rubric feedback and online policy optimization to follow-up work.

\bibliographystyle{unsrt}
\bibliography{ref}

@String{Computer = "{IEEE} Computer" }

@String{Chelsea = "Chelsea" }

@String{Springer = "Springer-Verlag" }

@inproceedings{dpo,
  author       = {Rafael Rafailov and
                  Archit Sharma and
                  Eric Mitchell and
                  Christopher D. Manning and
                  Stefano Ermon and
                  Chelsea Finn},
  editor       = {Alice Oh and
                  Tristan Naumann and
                  Amir Globerson and
                  Kate Saenko and
                  Moritz Hardt and
                  Sergey Levine},
  title        = {Direct Preference Optimization: Your Language Model is Secretly a
                  Reward Model},
  booktitle    = {Advances in Neural Information Processing Systems 36: Annual Conference
                  on Neural Information Processing Systems 2023, NeurIPS 2023, New Orleans,
                  LA, USA, December 10 - 16, 2023},
  year         = {2023},
  url          = {http://papers.nips.cc/paper\_files/paper/2023/hash/a85b405ed65c6477a4fe8302b5e06ce7-Abstract-Conference.html},
  bibsource    = {dblp computer science bibliography, https://dblp.org}
}

@article{bradley-terry,
  title={Rank analysis of incomplete block designs: I. the method of paired comparisons},
  author={Bradley, Ralph Allan and Terry, Milton E},
  journal={Biometrika},
  volume={39},
  number={3/4},
  pages={324--345},
  year={1952},
  publisher={JSTOR}
}

@article{smaug,
  author       = {Arka Pal and
                  Deep Karkhanis and
                  Samuel Dooley and
                  Manley Roberts and
                  Siddartha Naidu and
                  Colin White},
  title        = {Smaug: Fixing Failure Modes of Preference Optimisation with DPO-Positive},
  journal      = {CoRR},
  volume       = {abs/2402.13228},
  year         = {2024},
  url          = {https://doi.org/10.48550/arXiv.2402.13228},
  doi          = {10.48550/ARXIV.2402.13228},
  eprinttype    = {arXiv},
  eprint       = {2402.13228},
  bibsource    = {dblp computer science bibliography, https://dblp.org}
}

@inproceedings{length-regularized-dpo,
  author       = {Ryan Park and
                  Rafael Rafailov and
                  Stefano Ermon and
                  Chelsea Finn},
  editor       = {Lun{-}Wei Ku and
                  Andre Martins and
                  Vivek Srikumar},
  title        = {Disentangling Length from Quality in Direct Preference Optimization},
  booktitle    = {Findings of the Association for Computational Linguistics, {ACL} 2024,
                  Bangkok, Thailand and virtual meeting, August 11-16, 2024},
  series       = {Findings of {ACL}},
  volume       = {{ACL} 2024},
  pages        = {4998--5017},
  publisher    = {Association for Computational Linguistics},
  year         = {2024},
  url          = {https://doi.org/10.18653/v1/2024.findings-acl.297},
  doi          = {10.18653/V1/2024.FINDINGS-ACL.297},
  bibsource    = {dblp computer science bibliography, https://dblp.org}
}

@article{texthawk2,
  author       = {Ya{-}Qi Yu and
                  Minghui Liao and
                  Jiwen Zhang and
                  Jihao Wu},
  title        = {TextHawk2: {A} Large Vision-Language Model Excels in Bilingual {OCR}
                  and Grounding with 16x Fewer Tokens},
  journal      = {CoRR},
  volume       = {abs/2410.05261},
  year         = {2024},
  url          = {https://doi.org/10.48550/arXiv.2410.05261},
  doi          = {10.48550/ARXIV.2410.05261},
  eprinttype   = {arXiv},
  eprint       = {2410.05261},
  bibsource    = {dblp computer science bibliography, https://dblp.org}
}

@article{qwen2.5-vl,
  author       = {Shuai Bai and
                  Keqin Chen and
                  Xuejing Liu and
                  Jialin Wang and
                  Wenbin Ge and
                  Sibo Song and
                  Kai Dang and
                  Peng Wang and
                  Shijie Wang and
                  Jun Tang and
                  Humen Zhong and
                  Yuanzhi Zhu and
                  Ming{-}Hsuan Yang and
                  Zhaohai Li and
                  Jianqiang Wan and
                  Pengfei Wang and
                  Wei Ding and
                  Zheren Fu and
                  Yiheng Xu and
                  Jiabo Ye and
                  Xi Zhang and
                  Tianbao Xie and
                  Zesen Cheng and
                  Hang Zhang and
                  Zhibo Yang and
                  Haiyang Xu and
                  Junyang Lin},
  title        = {Qwen2.5-VL Technical Report},
  journal      = {CoRR},
  volume       = {abs/2502.13923},
  year         = {2025},
  url          = {https://doi.org/10.48550/arXiv.2502.13923},
  doi          = {10.48550/ARXIV.2502.13923},
  eprinttype    = {arXiv},
  eprint       = {2502.13923},
  bibsource    = {dblp computer science bibliography, https://dblp.org}
}

@article{mimo-vl,
  author       = {Zihao Yue and
                  Zhenru Lin and
                  Yifan Song and
                  Weikun Wang and
                  Shuhuai Ren and
                  Shuhao Gu and
                  Shicheng Li and
                  Peidian Li and
                  Liang Zhao and
                  Lei Li and
                  Kainan Bao and
                  Hao Tian and
                  Hailin Zhang and
                  Xiao{-}Gang Wang and
                  Dawei Zhu and
                  Cici and
                  Chenhong He and
                  Bowen Ye and
                  Bowen Shen and
                  Zihan Zhang and
                  Zihan Jiang and
                  Zhixian Zheng and
                  Zhichao Song and
                  Zhenbo Luo and
                  Yue Yu and
                  Yudong Wang and
                  Yuanyuan Tian and
                  Yu Tu and
                  Yihan Yan and
                  Yi Huang and
                  Xu Wang and
                  Xinzhe Xu and
                  Xingchen Song and
                  Xing Zhang and
                  Xing Yong and
                  Xin Zhang and
                  Xiangwei Deng and
                  Wenyu Yang and
                  Wenhan Ma and
                  Weiwei Lv and
                  Weiji Zhuang and
                  Wei Liu and
                  Sirui Deng and
                  Shuo Liu and
                  Shimao Chen and
                  Shihua Yu and
                  Shaohui Liu and
                  Shande Wang and
                  Rui Ma and
                  Qiantong Wang and
                  Peng Wang and
                  Nuo Chen and
                  Menghang Zhu and
                  Kangyang Zhou and
                  Kang Zhou and
                  Kai Fang and
                  Jun Shi and
                  Jinhao Dong and
                  Jiebao Xiao and
                  Jiaming Xu and
                  Huaqiu Liu and
                  Hongshen Xu and
                  Heng Qu and
                  Haochen Zhao and
                  Hanglong Lv and
                  Guoan Wang and
                  Duo Zhang and
                  Dong Zhang and
                  Di Zhang and
                  Chong Ma and
                  Chang Liu and
                  Can Cai and
                  Bingquan Xia},
  title        = {MiMo-VL Technical Report},
  journal      = {CoRR},
  volume       = {abs/2506.03569},
  year         = {2025},
  url          = {https://doi.org/10.48550/arXiv.2506.03569},
  doi          = {10.48550/ARXIV.2506.03569},
  eprinttype   = {arXiv},
  eprint       = {2506.03569},
  bibsource    = {dblp computer science bibliography, https://dblp.org}
}

@article{internvl-3.5,
  author       = {Weiyun Wang and
                  Zhangwei Gao and
                  Lixin Gu and
                  Hengjun Pu and
                  Long Cui and
                  Xingguang Wei and
                  Zhaoyang Liu and
                  Linglin Jing and
                  Shenglong Ye and
                  Jie Shao and
                  Zhaokai Wang and
                  Zhe Chen and
                  Hongjie Zhang and
                  Ganlin Yang and
                  Haomin Wang and
                  Qi Wei and
                  Jinhui Yin and
                  Wenhao Li and
                  Erfei Cui and
                  Guanzhou Chen and
                  Zichen Ding and
                  Changyao Tian and
                  Zhenyu Wu and
                  JingJing Xie and
                  Zehao Li and
                  Bowen Yang and
                  Yuchen Duan and
                  Xuehui Wang and
                  Zhi Hou and
                  Haoran Hao and
                  Tianyi Zhang and
                  Songze Li and
                  Xiangyu Zhao and
                  Haodong Duan and
                  Nianchen Deng and
                  Bin Fu and
                  Yinan He and
                  Yi Wang and
                  Conghui He and
                  Botian Shi and
                  Junjun He and
                  Yingtong Xiong and
                  Han Lv and
                  Lijun Wu and
                  Wenqi Shao and
                  Kaipeng Zhang and
                  Huipeng Deng and
                  Biqing Qi and
                  Jiaye Ge and
                  Qipeng Guo and
                  Wenwei Zhang and
                  Songyang Zhang and
                  Maosong Cao and
                  Junyao Lin and
                  Kexian Tang and
                  Jianfei Gao and
                  Haian Huang and
                  Yuzhe Gu and
                  Chengqi Lyu and
                  Huanze Tang and
                  Rui Wang and
                  Haijun Lv and
                  Wanli Ouyang and
                  Limin Wang and
                  Min Dou and
                  Xizhou Zhu and
                  Tong Lu and
                  Dahua Lin and
                  Jifeng Dai and
                  Weijie Su and
                  Bowen Zhou and
                  Kai Chen and
                  Yu Qiao and
                  Wenhai Wang and
                  Gen Luo},
  title        = {InternVL3.5: Advancing Open-Source Multimodal Models in Versatility,
                  Reasoning, and Efficiency},
  journal      = {CoRR},
  volume       = {abs/2508.18265},
  year         = {2025},
  url          = {https://doi.org/10.48550/arXiv.2508.18265},
  doi          = {10.48550/ARXIV.2508.18265},
  eprinttype    = {arXiv},
  eprint       = {2508.18265},
  bibsource    = {dblp computer science bibliography, https://dblp.org}
}

@article{keye-vl-1.5,
  author       = {Biao Yang and
                  Bin Wen and
                  Boyang Ding and
                  Changyi Liu and
                  Chenglong Chu and
                  Chengru Song and
                  Chongling Rao and
                  Chuan Yi and
                  Da Li and
                  Dunju Zang and
                  Fan Yang and
                  Guorui Zhou and
                  Guowang Zhang and
                  Han Shen and
                  Hao Peng and
                  Haojie Ding and
                  Hao Wang and
                  Haonan Fan and
                  Hengrui Ju and
                  Jiaming Huang and
                  Jiangxia Cao and
                  Jiankang Chen and
                  Jingyun Hua and
                  Kaibing Chen and
                  Kaiyu Jiang and
                  Kaiyu Tang and
                  Kun Gai and
                  Muhao Wei and
                  Qiang Wang and
                  Ruitao Wang and
                  Sen Na and
                  Shengnan Zhang and
                  Siyang Mao and
                  Sui Huang and
                  Tianke Zhang and
                  Tingting Gao and
                  Wei Chen and
                  Wei Yuan and
                  Xiangyu Wu and
                  Xiao Hu and
                  Xingyu Lu and
                  Yifan Zhang and
                  Yiping Yang and
                  Yulong Chen and
                  Zeyi Lu and
                  Zhenhua Wu and
                  Zhixin Ling and
                  Zhuoran Yang and
                  Ziming Li and
                  Di Xu and
                  Haixuan Gao and
                  Hang Li and
                  Jing Wang and
                  Lejian Ren and
                  Qigen Hu and
                  Qianqian Wang and
                  Shiyao Wang and
                  Xinchen Luo and
                  Yan Li and
                  Yuhang Hu and
                  Zixing Zhang},
  title        = {Kwai Keye-VL 1.5 Technical Report},
  journal      = {CoRR},
  volume       = {abs/2509.01563},
  year         = {2025},
  url          = {https://doi.org/10.48550/arXiv.2509.01563},
  doi          = {10.48550/ARXIV.2509.01563},
  eprinttype    = {arXiv},
  eprint       = {2509.01563},
  bibsource    = {dblp computer science bibliography, https://dblp.org}
}

@article{qwen3-vl,
  author       = {Qwen Team},
  title        = {Qwen3-VL Technical Report},
  journal      = {CoRR},
  volume       = {abs/2511.21631},
  year         = {2025},
  url          = {https://doi.org/10.48550/arXiv.2511.21631},
  doi          = {10.48550/ARXIV.2511.21631},
  eprinttype    = {arXiv},
  eprint       = {2511.21631},
  bibsource    = {dblp computer science bibliography, https://dblp.org}
}

@inproceedings{ha-dpo,
  author       = {Zhiyuan Zhao and
                  Bin Wang and
                  Linke Ouyang and
                  Xiaoyi Dong and
                  Jiaqi Wang and
                  Conghui He},
  title        = {Beyond Multimodal Hallucinations: Enhancing LVLMs through Hallucination-Aware
                  Direct Preference Optimization},
  booktitle    = {{IEEE} International Conference on Multimedia and Expo, {ICME} 2025,
                  Nantes, France, June 30 - July 4, 2025},
  pages        = {1--6},
  publisher    = {{IEEE}},
  year         = {2025},
  url          = {https://doi.org/10.1109/ICME59968.2025.11209377},
  doi          = {10.1109/ICME59968.2025.11209377},
  bibsource    = {dblp computer science bibliography, https://dblp.org}
}

@inproceedings{bpo,
  author       = {Renjie Pi and
                  Tianyang Han and
                  Wei Xiong and
                  Jipeng Zhang and
                  Runtao Liu and
                  Rui Pan and
                  Tong Zhang},
  editor       = {Ales Leonardis and
                  Elisa Ricci and
                  Stefan Roth and
                  Olga Russakovsky and
                  Torsten Sattler and
                  G{\"{u}}l Varol},
  title        = {Strengthening Multimodal Large Language Model with Bootstrapped Preference
                  Optimization},
  booktitle    = {Computer Vision - {ECCV} 2024 - 18th European Conference, Milan, Italy,
                  September 29-October 4, 2024, Proceedings, Part {XXXIII}},
  series       = {Lecture Notes in Computer Science},
  volume       = {15091},
  pages        = {382--398},
  publisher    = {Springer},
  year         = {2024},
  url          = {https://doi.org/10.1007/978-3-031-73414-4\_22},
  doi          = {10.1007/978-3-031-73414-4\_22},
  bibsource    = {dblp computer science bibliography, https://dblp.org}
}

@inproceedings{stic,
  author       = {Yihe Deng and
                  Pan Lu and
                  Fan Yin and
                  Ziniu Hu and
                  Sheng Shen and
                  Quanquan Gu and
                  James Y. Zou and
                  Kai{-}Wei Chang and
                  Wei Wang},
  editor       = {Amir Globersons and
                  Lester Mackey and
                  Danielle Belgrave and
                  Angela Fan and
                  Ulrich Paquet and
                  Jakub M. Tomczak and
                  Cheng Zhang},
  title        = {Enhancing Large Vision Language Models with Self-Training on Image
                  Comprehension},
  booktitle    = {Advances in Neural Information Processing Systems 38: Annual Conference
                  on Neural Information Processing Systems 2024, NeurIPS 2024, Vancouver,
                  BC, Canada, December 10 - 15, 2024},
  year         = {2024},
  url          = {http://papers.nips.cc/paper\_files/paper/2024/hash/ed45d6a03de84cc650cae0655f699356-Abstract-Conference.html},
  bibsource    = {dblp computer science bibliography, https://dblp.org}
}

@inproceedings{seva,
  author       = {Ke Zhu and
                  Liang Zhao and
                  Zheng Ge and
                  Xiangyu Zhang},
  editor       = {Jianfei Cai and
                  Mohan S. Kankanhalli and
                  Balakrishnan Prabhakaran and
                  Susanne Boll and
                  Ramanathan Subramanian and
                  Liang Zheng and
                  Vivek K. Singh and
                  Pablo C{\'{e}}sar and
                  Lexing Xie and
                  Dong Xu},
  title        = {Self-Supervised Visual Preference Alignment},
  booktitle    = {Proceedings of the 32nd {ACM} International Conference on Multimedia,
                  {MM} 2024, Melbourne, VIC, Australia, 28 October 2024 - 1 November
                  2024},
  pages        = {291--300},
  publisher    = {{ACM}},
  year         = {2024},
  url          = {https://doi.org/10.1145/3664647.3680993},
  doi          = {10.1145/3664647.3680993},
  bibsource    = {dblp computer science bibliography, https://dblp.org}
}

@inproceedings{rlaif-v,
  author       = {Tianyu Yu and
                  Haoye Zhang and
                  Qiming Li and
                  Qixin Xu and
                  Yuan Yao and
                  Da Chen and
                  Xiaoman Lu and
                  Ganqu Cui and
                  Yunkai Dang and
                  Taiwen He and
                  Xiaocheng Feng and
                  Jun Song and
                  Bo Zheng and
                  Zhiyuan Liu and
                  Tat{-}Seng Chua and
                  Maosong Sun},
  title        = {{RLAIF-V:} Open-Source {AI} Feedback Leads to Super {GPT-4V} Trustworthiness},
  booktitle    = {{IEEE/CVF} Conference on Computer Vision and Pattern Recognition,
                  {CVPR} 2025, Nashville, TN, USA, June 11-15, 2025},
  pages        = {19985--19995},
  publisher    = {Computer Vision Foundation / {IEEE}},
  year         = {2025},
  url          = {https://openaccess.thecvf.com/content/CVPR2025/html/Yu\_RLAIF-V\_Open-Source\_AI\_Feedback\_Leads\_to\_Super\_GPT-4V\_Trustworthiness\_CVPR\_2025\_paper.html},
  doi          = {10.1109/CVPR52734.2025.01861},
  bibsource    = {dblp computer science bibliography, https://dblp.org}
}

@article{self-improve,
  author       = {Shijian Deng and
                  Wentian Zhao and
                  Yu{-}Jhe Li and
                  Kun Wan and
                  Daniel Miranda and
                  Ajinkya Kale and
                  Yapeng Tian},
  title        = {Efficient Self-Improvement in Multimodal Large Language Models: {A}
                  Model-Level Judge-Free Approach},
  journal      = {CoRR},
  volume       = {abs/2411.17760},
  year         = {2024},
  url          = {https://doi.org/10.48550/arXiv.2411.17760},
  doi          = {10.48550/ARXIV.2411.17760},
  eprinttype    = {arXiv},
  eprint       = {2411.17760},
  bibsource    = {dblp computer science bibliography, https://dblp.org}
}

@article{mpo,
  author       = {Weiyun Wang and
                  Zhe Chen and
                  Wenhai Wang and
                  Yue Cao and
                  Yangzhou Liu and
                  Zhangwei Gao and
                  Jinguo Zhu and
                  Xizhou Zhu and
                  Lewei Lu and
                  Yu Qiao and
                  Jifeng Dai},
  title        = {Enhancing the Reasoning Ability of Multimodal Large Language Models
                  via Mixed Preference Optimization},
  journal      = {CoRR},
  volume       = {abs/2411.10442},
  year         = {2024},
  url          = {https://doi.org/10.48550/arXiv.2411.10442},
  doi          = {10.48550/ARXIV.2411.10442},
  eprinttype    = {arXiv},
  eprint       = {2411.10442},
  bibsource    = {dblp computer science bibliography, https://dblp.org}
}

@article{povid,
  author       = {Yiyang Zhou and
                  Chenhang Cui and
                  Rafael Rafailov and
                  Chelsea Finn and
                  Huaxiu Yao},
  title        = {Aligning Modalities in Vision Large Language Models via Preference
                  Fine-tuning},
  journal      = {CoRR},
  volume       = {abs/2402.11411},
  year         = {2024},
  url          = {https://doi.org/10.48550/arXiv.2402.11411},
  doi          = {10.48550/ARXIV.2402.11411},
  eprinttype    = {arXiv},
  eprint       = {2402.11411},
  bibsource    = {dblp computer science bibliography, https://dblp.org}
}

@inproceedings{mdpo,
  author       = {Fei Wang and
                  Wenxuan Zhou and
                  James Y. Huang and
                  Nan Xu and
                  Sheng Zhang and
                  Hoifung Poon and
                  Muhao Chen},
  editor       = {Yaser Al{-}Onaizan and
                  Mohit Bansal and
                  Yun{-}Nung Chen},
  title        = {mDPO: Conditional Preference Optimization for Multimodal Large Language
                  Models},
  booktitle    = {Proceedings of the 2024 Conference on Empirical Methods in Natural
                  Language Processing, {EMNLP} 2024, Miami, FL, USA, November 12-16,
                  2024},
  pages        = {8078--8088},
  publisher    = {Association for Computational Linguistics},
  year         = {2024},
  url          = {https://doi.org/10.18653/v1/2024.emnlp-main.460},
  doi          = {10.18653/V1/2024.EMNLP-MAIN.460},
  bibsource    = {dblp computer science bibliography, https://dblp.org}
}

@inproceedings{mfdpo,
  author       = {Songtao Jiang and
                  Yan Zhang and
                  Ruizhe Chen and
                  Tianxiang Hu and
                  Yeying Jin and
                  Qinglin He and
                  Yang Feng and
                  Jian Wu and
                  Zuozhu Liu},
  title        = {Modality-Fair Preference Optimization for Trustworthy {MLLM} Alignment},
  booktitle    = {Proceedings of the Thirty-Fourth International Joint Conference on
                  Artificial Intelligence, {IJCAI} 2025, Montreal, Canada, August 16-22,
                  2025},
  pages        = {403--411},
  publisher    = {ijcai.org},
  year         = {2025},
  url          = {https://doi.org/10.24963/ijcai.2025/46},
  doi          = {10.24963/IJCAI.2025/46},
  bibsource    = {dblp computer science bibliography, https://dblp.org}
}

@inproceedings{v-dpo,
  author       = {Yuxi Xie and
                  Guanzhen Li and
                  Xiao Xu and
                  Min{-}Yen Kan},
  editor       = {Yaser Al{-}Onaizan and
                  Mohit Bansal and
                  Yun{-}Nung Chen},
  title        = {{V-DPO:} Mitigating Hallucination in Large Vision Language Models
                  via Vision-Guided Direct Preference Optimization},
  booktitle    = {Findings of the Association for Computational Linguistics: {EMNLP}
                  2024, Miami, Florida, USA, November 12-16, 2024},
  series       = {Findings of {ACL}},
  volume       = {{EMNLP} 2024},
  pages        = {13258--13273},
  publisher    = {Association for Computational Linguistics},
  year         = {2024},
  url          = {https://doi.org/10.18653/v1/2024.findings-emnlp.775},
  doi          = {10.18653/V1/2024.FINDINGS-EMNLP.775},
  bibsource    = {dblp computer science bibliography, https://dblp.org}
}

@inproceedings{re-align,
  author       = {Shuo Xing and
                  Peiran Li and
                  Yuping Wang and
                  Ruizheng Bai and
                  Yueqi Wang and
                  Chan{-}Wei Hu and
                  Chengxuan Qian and
                  Huaxiu Yao and
                  Zhengzhong Tu},
  editor       = {Christos Christodoulopoulos and
                  Tanmoy Chakraborty and
                  Carolyn Rose and
                  Violet Peng},
  title        = {Re-Align: Aligning Vision Language Models via Retrieval-Augmented
                  Direct Preference Optimization},
  booktitle    = {Proceedings of the 2025 Conference on Empirical Methods in Natural
                  Language Processing, {EMNLP} 2025, Suzhou, China, November 4-9, 2025},
  pages        = {2379--2397},
  publisher    = {Association for Computational Linguistics},
  year         = {2025},
  url          = {https://doi.org/10.18653/v1/2025.emnlp-main.121},
  doi          = {10.18653/V1/2025.EMNLP-MAIN.121},
  bibsource    = {dblp computer science bibliography, https://dblp.org}
}

@inproceedings{vilp,
  author       = {Tiange Luo and
                  Ang Cao and
                  Gunhee Lee and
                  Justin Johnson and
                  Honglak Lee},
  editor       = {Aarti Singh and
                  Maryam Fazel and
                  Daniel Hsu and
                  Simon Lacoste{-}Julien and
                  Felix Berkenkamp and
                  Tegan Maharaj and
                  Kiri Wagstaff and
                  Jerry Zhu},
  title        = {Probing Visual Language Priors in VLMs},
  booktitle    = {Forty-second International Conference on Machine Learning, {ICML}
                  2025, Vancouver, BC, Canada, July 13-19, 2025},
  series       = {Proceedings of Machine Learning Research},
  volume       = {267},
  publisher    = {{PMLR} / OpenReview.net},
  year         = {2025},
  url          = {https://proceedings.mlr.press/v267/luo25b.html},
  bibsource    = {dblp computer science bibliography, https://dblp.org}
}

@article{symmpo,
  author       = {Wenqi Liu and
                  Xuemeng Song and
                  Jiaxi Li and
                  Yinwei Wei and
                  Na Zheng and
                  Jianhua Yin and
                  Liqiang Nie},
  title        = {Mitigating Hallucination Through Theory-Consistent Symmetric Multimodal
                  Preference Optimization},
  journal      = {CoRR},
  volume       = {abs/2506.11712},
  year         = {2025},
  url          = {https://doi.org/10.48550/arXiv.2506.11712},
  doi          = {10.48550/ARXIV.2506.11712},
  eprinttype    = {arXiv},
  eprint       = {2506.11712},
  bibsource    = {dblp computer science bibliography, https://dblp.org}
}

@inproceedings{opa-dpo,
  author       = {Zhihe Yang and
                  Xufang Luo and
                  Dongqi Han and
                  Yunjian Xu and
                  Dongsheng Li},
  title        = {Mitigating Hallucinations in Large Vision-Language Models via {DPO:}
                  On-Policy Data Hold the Key},
  booktitle    = {{IEEE/CVF} Conference on Computer Vision and Pattern Recognition,
                  {CVPR} 2025, Nashville, TN, USA, June 11-15, 2025},
  pages        = {10610--10620},
  publisher    = {Computer Vision Foundation / {IEEE}},
  year         = {2025},
  url          = {https://openaccess.thecvf.com/content/CVPR2025/html/Yang\_Mitigating\_Hallucinations\_in\_Large\_Vision-Language\_Models\_via\_DPO\_On-Policy\_Data\_CVPR\_2025\_paper.html},
  doi          = {10.1109/CVPR52734.2025.00992},
  bibsource    = {dblp computer science bibliography, https://dblp.org}
}

@article{ovip,
  author       = {Shujun Liu and
                  Siyuan Wang and
                  Zejun Li and
                  Jianxiang Wang and
                  Cheng Zeng and
                  Zhongyu Wei},
  title        = {OViP: Online Vision-Language Preference Learning},
  journal      = {CoRR},
  volume       = {abs/2505.15963},
  year         = {2025},
  url          = {https://doi.org/10.48550/arXiv.2505.15963},
  doi          = {10.48550/ARXIV.2505.15963},
  eprinttype    = {arXiv},
  eprint       = {2505.15963},
  bibsource    = {dblp computer science bibliography, https://dblp.org}
}

@article{on-policy-iterative-dpo,
  author       = {Chengzhi Yu and
                  Yifan Xu and
                  Yifan Chen and
                  Wenyi Zhang},
  title        = {Optimizing LVLMs with On-Policy Data for Effective Hallucination Mitigation},
  journal      = {CoRR},
  volume       = {abs/2512.00706},
  year         = {2025},
  url          = {https://doi.org/10.48550/arXiv.2512.00706},
  doi          = {10.48550/ARXIV.2512.00706},
  eprinttype    = {arXiv},
  eprint       = {2512.00706},
  bibsource    = {dblp computer science bibliography, https://dblp.org}
}

@inproceedings{sima,
  author       = {Xiyao Wang and
                  Jiuhai Chen and
                  Zhaoyang Wang and
                  Yuhang Zhou and
                  Yiyang Zhou and
                  Huaxiu Yao and
                  Tianyi Zhou and
                  Tom Goldstein and
                  Parminder Bhatia and
                  Taha A. Kass{-}Hout and
                  Furong Huang and
                  Cao Xiao},
  editor       = {Luis Chiruzzo and
                  Alan Ritter and
                  Lu Wang},
  title        = {Enhancing Visual-Language Modality Alignment in Large Vision Language
                  Models via Self-Improvement},
  booktitle    = {Findings of the Association for Computational Linguistics: {NAACL}
                  2025, Albuquerque, New Mexico, USA, April 29 - May 4, 2025},
  series       = {Findings of {ACL}},
  volume       = {{NAACL} 2025},
  pages        = {268--282},
  publisher    = {Association for Computational Linguistics},
  year         = {2025},
  url          = {https://doi.org/10.18653/v1/2025.findings-naacl.15},
  doi          = {10.18653/V1/2025.FINDINGS-NAACL.15},
  bibsource    = {dblp computer science bibliography, https://dblp.org}
}

@inproceedings{llava-rlhf,
  author       = {Zhiqing Sun and
                  Sheng Shen and
                  Shengcao Cao and
                  Haotian Liu and
                  Chunyuan Li and
                  Yikang Shen and
                  Chuang Gan and
                  Liangyan Gui and
                  Yu{-}Xiong Wang and
                  Yiming Yang and
                  Kurt Keutzer and
                  Trevor Darrell},
  editor       = {Lun{-}Wei Ku and
                  Andre Martins and
                  Vivek Srikumar},
  title        = {Aligning Large Multimodal Models with Factually Augmented {RLHF}},
  booktitle    = {Findings of the Association for Computational Linguistics, {ACL} 2024,
                  Bangkok, Thailand and virtual meeting, August 11-16, 2024},
  series       = {Findings of {ACL}},
  volume       = {{ACL} 2024},
  pages        = {13088--13110},
  publisher    = {Association for Computational Linguistics},
  year         = {2024},
  url          = {https://doi.org/10.18653/v1/2024.findings-acl.775},
  doi          = {10.18653/V1/2024.FINDINGS-ACL.775},
  bibsource    = {dblp computer science bibliography, https://dblp.org}
}

@inproceedings{rlhf-v,
  author       = {Tianyu Yu and
                  Yuan Yao and
                  Haoye Zhang and
                  Taiwen He and
                  Yifeng Han and
                  Ganqu Cui and
                  Jinyi Hu and
                  Zhiyuan Liu and
                  Hai{-}Tao Zheng and
                  Maosong Sun},
  title        = {{RLHF-V:} Towards Trustworthy MLLMs via Behavior Alignment from Fine-Grained
                  Correctional Human Feedback},
  booktitle    = {{IEEE/CVF} Conference on Computer Vision and Pattern Recognition,
                  {CVPR} 2024, Seattle, WA, USA, June 16-22, 2024},
  pages        = {13807--13816},
  publisher    = {{IEEE}},
  year         = {2024},
  url          = {https://doi.org/10.1109/CVPR52733.2024.01310},
  doi          = {10.1109/CVPR52733.2024.01310},
  bibsource    = {dblp computer science bibliography, https://dblp.org}
}

@inproceedings{wildvision,
  author       = {Yujie Lu and
                  Dongfu Jiang and
                  Wenhu Chen and
                  William Yang Wang and
                  Yejin Choi and
                  Bill Yuchen Lin},
  editor       = {Amir Globersons and
                  Lester Mackey and
                  Danielle Belgrave and
                  Angela Fan and
                  Ulrich Paquet and
                  Jakub M. Tomczak and
                  Cheng Zhang},
  title        = {WildVision: Evaluating Vision-Language Models in the Wild with Human
                  Preferences},
  booktitle    = {Advances in Neural Information Processing Systems 38: Annual Conference
                  on Neural Information Processing Systems 2024, NeurIPS 2024, Vancouver,
                  BC, Canada, December 10 - 15, 2024},
  year         = {2024},
  url          = {http://papers.nips.cc/paper\_files/paper/2024/hash/563991b5c8b45fe75bea42db738223b2-Abstract-Datasets\_and\_Benchmarks\_Track.html},
  bibsource    = {dblp computer science bibliography, https://dblp.org}
}

@article{silkie,
  author       = {Lei Li and
                  Zhihui Xie and
                  Mukai Li and
                  Shunian Chen and
                  Peiyi Wang and
                  Liang Chen and
                  Yazheng Yang and
                  Benyou Wang and
                  Lingpeng Kong},
  title        = {Silkie: Preference Distillation for Large Visual Language Models},
  journal      = {CoRR},
  volume       = {abs/2312.10665},
  year         = {2023},
  url          = {https://doi.org/10.48550/arXiv.2312.10665},
  doi          = {10.48550/ARXIV.2312.10665},
  eprinttype    = {arXiv},
  eprint       = {2312.10665},
  bibsource    = {dblp computer science bibliography, https://dblp.org}
}

@inproceedings{mia-dpo,
  author       = {Ziyu Liu and
                  Yuhang Zang and
                  Xiaoyi Dong and
                  Pan Zhang and
                  Yuhang Cao and
                  Haodong Duan and
                  Conghui He and
                  Yuanjun Xiong and
                  Dahua Lin and
                  Jiaqi Wang},
  title        = {{MIA-DPO:} Multi-Image Augmented Direct Preference Optimization For
                  Large Vision-Language Models},
  booktitle    = {The Thirteenth International Conference on Learning Representations,
                  {ICLR} 2025, Singapore, April 24-28, 2025},
  publisher    = {OpenReview.net},
  year         = {2025},
  url          = {https://openreview.net/forum?id=f7WBRSuf9l},
  bibsource    = {dblp computer science bibliography, https://dblp.org}
}

@inproceedings{mm-rlhf,
  author       = {Yifan Zhang and
                  Tao Yu and
                  Haochen Tian and
                  Chaoyou Fu and
                  Peiyan Li and
                  Jianshu Zeng and
                  Wulin Xie and
                  Yang Shi and
                  Huanyu Zhang and
                  Junkang Wu and
                  Xue Wang and
                  Yibo Hu and
                  Bin Wen and
                  Tingting Gao and
                  Zhang Zhang and
                  Fan Yang and
                  Di Zhang and
                  Liang Wang and
                  Rong Jin},
  editor       = {Aarti Singh and
                  Maryam Fazel and
                  Daniel Hsu and
                  Simon Lacoste{-}Julien and
                  Felix Berkenkamp and
                  Tegan Maharaj and
                  Kiri Wagstaff and
                  Jerry Zhu},
  title        = {{MM-RLHF:} The Next Step Forward in Multimodal {LLM} Alignment},
  booktitle    = {Forty-second International Conference on Machine Learning, {ICML}
                  2025, Vancouver, BC, Canada, July 13-19, 2025},
  series       = {Proceedings of Machine Learning Research},
  volume       = {267},
  publisher    = {{PMLR} / OpenReview.net},
  year         = {2025},
  url          = {https://proceedings.mlr.press/v267/zhang25cs.html},
  bibsource    = {dblp computer science bibliography, https://dblp.org}
}

@inproceedings{prometheus-vision,
  author       = {Seongyun Lee and
                  Seungone Kim and
                  Sue Hyun Park and
                  Geewook Kim and
                  Minjoon Seo},
  editor       = {Lun{-}Wei Ku and
                  Andre Martins and
                  Vivek Srikumar},
  title        = {Prometheus-Vision: Vision-Language Model as a Judge for Fine-Grained
                  Evaluation},
  booktitle    = {Findings of the Association for Computational Linguistics, {ACL} 2024,
                  Bangkok, Thailand and virtual meeting, August 11-16, 2024},
  series       = {Findings of {ACL}},
  volume       = {{ACL} 2024},
  pages        = {11286--11315},
  publisher    = {Association for Computational Linguistics},
  year         = {2024},
  url          = {https://doi.org/10.18653/v1/2024.findings-acl.672},
  doi          = {10.18653/V1/2024.FINDINGS-ACL.672},
  bibsource    = {dblp computer science bibliography, https://dblp.org}
}

@inproceedings{llava-critic,
  author       = {Tianyi Xiong and
                  Xiyao Wang and
                  Dong Guo and
                  Qinghao Ye and
                  Haoqi Fan and
                  Quanquan Gu and
                  Heng Huang and
                  Chunyuan Li},
  title        = {LLaVA-Critic: Learning to Evaluate Multimodal Models},
  booktitle    = {{IEEE/CVF} Conference on Computer Vision and Pattern Recognition,
                  {CVPR} 2025, Nashville, TN, USA, June 11-15, 2025},
  pages        = {13618--13628},
  publisher    = {Computer Vision Foundation / {IEEE}},
  year         = {2025},
  url          = {https://openaccess.thecvf.com/content/CVPR2025/html/Xiong\_LLaVA-Critic\_Learning\_to\_Evaluate\_Multimodal\_Models\_CVPR\_2025\_paper.html},
  doi          = {10.1109/CVPR52734.2025.01271},
  bibsource    = {dblp computer science bibliography, https://dblp.org}
}

@inproceedings{critic-v,
  author       = {Di Zhang and
                  Jingdi Lei and
                  Junxian Li and
                  Xunzhi Wang and
                  Yujie Liu and
                  Zonglin Yang and
                  Jiatong Li and
                  Weida Wang and
                  Suorong Yang and
                  Jianbo Wu and
                  Peng Ye and
                  Wanli Ouyang and
                  Dongzhan Zhou},
  title        = {Critic-V: {VLM} Critics Help Catch {VLM} Errors in Multimodal Reasoning},
  booktitle    = {{IEEE/CVF} Conference on Computer Vision and Pattern Recognition,
                  {CVPR} 2025, Nashville, TN, USA, June 11-15, 2025},
  pages        = {9050--9061},
  publisher    = {Computer Vision Foundation / {IEEE}},
  year         = {2025},
  url          = {https://openaccess.thecvf.com/content/CVPR2025/html/Zhang\_Critic-V\_VLM\_Critics\_Help\_Catch\_VLM\_Errors\_in\_Multimodal\_Reasoning\_CVPR\_2025\_paper.html},
  doi          = {10.1109/CVPR52734.2025.00846},
  bibsource    = {dblp computer science bibliography, https://dblp.org}
}

@inproceedings{ixc-2.5-reward,
  author       = {Yuhang Zang and
                  Xiaoyi Dong and
                  Pan Zhang and
                  Yuhang Cao and
                  Ziyu Liu and
                  Shengyuan Ding and
                  Shenxi Wu and
                  Yubo Ma and
                  Haodong Duan and
                  Wenwei Zhang and
                  Kai Chen and
                  Dahua Lin and
                  Jiaqi Wang},
  editor       = {Wanxiang Che and
                  Joyce Nabende and
                  Ekaterina Shutova and
                  Mohammad Taher Pilehvar},
  title        = {InternLM-XComposer2.5-Reward: {A} Simple Yet Effective Multi-Modal
                  Reward Model},
  booktitle    = {Findings of the Association for Computational Linguistics, {ACL} 2025,
                  Vienna, Austria, July 27 - August 1, 2025},
  series       = {Findings of {ACL}},
  volume       = {{ACL} 2025},
  pages        = {6547--6563},
  publisher    = {Association for Computational Linguistics},
  year         = {2025},
  url          = {https://aclanthology.org/2025.findings-acl.340/},
  bibsource    = {dblp computer science bibliography, https://dblp.org}
}

@inproceedings{carevl,
  author       = {Muzhi Dai and
                  Jiashuo Sun and
                  Zhiyuan Zhao and
                  Shixuan Liu and
                  Rui Li and
                  Junyu Gao and
                  Xuelong Li},
  editor       = {Cathal Gurrin and
                  Klaus Schoeffmann and
                  Min Zhang and
                  Luca Rossetto and
                  Stevan Rudinac and
                  Duc{-}Tien Dang{-}Nguyen and
                  Wen{-}Huang Cheng and
                  Phoebe Chen and
                  Jenny Benois{-}Pineau},
  title        = {From Captions to Rewards (CaReVL): Leveraging Large Language Model
                  Experts for Enhanced Reward Modeling in Large Vision-Language Models},
  booktitle    = {Proceedings of the 33rd {ACM} International Conference on Multimedia,
                  {MM} 2025, Dublin, Ireland, October 27-31, 2025},
  pages        = {4972--4981},
  publisher    = {{ACM}},
  year         = {2025},
  url          = {https://doi.org/10.1145/3746027.3755697},
  doi          = {10.1145/3746027.3755697},
  bibsource    = {dblp computer science bibliography, https://dblp.org}
}

@article{svip-reward,
  author       = {Minghe Gao and
                  Xuqi Liu and
                  Zhongqi Yue and
                  Yang Wu and
                  Shuang Chen and
                  Juncheng Li and
                  Siliang Tang and
                  Fei Wu and
                  Tat{-}Seng Chua and
                  Yueting Zhuang},
  title        = {Benchmarking Multimodal CoT Reward Model Stepwise by Visual Program},
  journal      = {CoRR},
  volume       = {abs/2504.06606},
  year         = {2025},
  url          = {https://doi.org/10.48550/arXiv.2504.06606},
  doi          = {10.48550/ARXIV.2504.06606},
  eprinttype    = {arXiv},
  eprint       = {2504.06606},
  bibsource    = {dblp computer science bibliography, https://dblp.org}
}

@article{skywork-vl-reward,
  author       = {Xiaokun Wang and
                  Peiyu Wang and
                  Jiangbo Pei and
                  Wei Shen and
                  Yi Peng and
                  Yunzhuo Hao and
                  Weijie Qiu and
                  Ai Jian and
                  Tianyidan Xie and
                  Xuchen Song and
                  Yang Liu and
                  Yahui Zhou},
  title        = {Skywork-VL Reward: An Effective Reward Model for Multimodal Understanding
                  and Reasoning},
  journal      = {CoRR},
  volume       = {abs/2505.07263},
  year         = {2025},
  url          = {https://doi.org/10.48550/arXiv.2505.07263},
  doi          = {10.48550/ARXIV.2505.07263},
  eprinttype    = {arXiv},
  eprint       = {2505.07263},
  bibsource    = {dblp computer science bibliography, https://dblp.org}
}

@inproceedings{llm-rubric,
  author       = {Helia Hashemi and
                  Jason Eisner and
                  Corby Rosset and
                  Benjamin Van Durme and
                  Chris Kedzie},
  editor       = {Lun{-}Wei Ku and
                  Andre Martins and
                  Vivek Srikumar},
  title        = {LLM-Rubric: {A} Multidimensional, Calibrated Approach to Automated
                  Evaluation of Natural Language Texts},
  booktitle    = {Proceedings of the 62nd Annual Meeting of the Association for Computational
                  Linguistics (Volume 1: Long Papers), {ACL} 2024, Bangkok, Thailand,
                  August 11-16, 2024},
  pages        = {13806--13834},
  publisher    = {Association for Computational Linguistics},
  year         = {2024},
  url          = {https://doi.org/10.18653/v1/2024.acl-long.745},
  doi          = {10.18653/V1/2024.ACL-LONG.745},
  bibsource    = {dblp computer science bibliography, https://dblp.org}
}

@inproceedings{paperbench,
  author       = {Giulio Starace and
                  Oliver Jaffe and
                  Dane Sherburn and
                  James Aung and
                  Jun Shern Chan and
                  Leon Maksin and
                  Rachel Dias and
                  Evan Mays and
                  Benjamin Kinsella and
                  Wyatt Thompson and
                  Johannes Heidecke and
                  Amelia Glaese and
                  Tejal Patwardhan},
  editor       = {Aarti Singh and
                  Maryam Fazel and
                  Daniel Hsu and
                  Simon Lacoste{-}Julien and
                  Felix Berkenkamp and
                  Tegan Maharaj and
                  Kiri Wagstaff and
                  Jerry Zhu},
  title        = {PaperBench: Evaluating AI's Ability to Replicate {AI} Research},
  booktitle    = {Forty-second International Conference on Machine Learning, {ICML}
                  2025, Vancouver, BC, Canada, July 13-19, 2025},
  series       = {Proceedings of Machine Learning Research},
  volume       = {267},
  publisher    = {{PMLR} / OpenReview.net},
  year         = {2025},
  url          = {https://proceedings.mlr.press/v267/starace25a.html},
  bibsource    = {dblp computer science bibliography, https://dblp.org}
}

@article{healthbench,
  author       = {Rahul K. Arora and
                  Jason Wei and
                  Rebecca Soskin Hicks and
                  Preston Bowman and
                  Joaquin Qui{\~{n}}onero Candela and
                  Foivos Tsimpourlas and
                  Michael Sharman and
                  Meghan Shah and
                  Andrea Vallone and
                  Alex Beutel and
                  Johannes Heidecke and
                  Karan Singhal},
  title        = {HealthBench: Evaluating Large Language Models Towards Improved Human
                  Health},
  journal      = {CoRR},
  volume       = {abs/2505.08775},
  year         = {2025},
  url          = {https://doi.org/10.48550/arXiv.2505.08775},
  doi          = {10.48550/ARXIV.2505.08775},
  eprinttype    = {arXiv},
  eprint       = {2505.08775},
  bibsource    = {dblp computer science bibliography, https://dblp.org}
}

@article{profbench,
  author       = {Zhilin Wang and
                  Jaehun Jung and
                  Ximing Lu and
                  Shizhe Diao and
                  Ellie Evans and
                  Jiaqi Zeng and
                  Pavlo Molchanov and
                  Yejin Choi and
                  Jan Kautz and
                  Yi Dong},
  title        = {ProfBench: Multi-Domain Rubrics requiring Professional Knowledge to
                  Answer and Judge},
  journal      = {CoRR},
  volume       = {abs/2510.18941},
  year         = {2025},
  url          = {https://doi.org/10.48550/arXiv.2510.18941},
  doi          = {10.48550/ARXIV.2510.18941},
  eprinttype    = {arXiv},
  eprint       = {2510.18941},
  bibsource    = {dblp computer science bibliography, https://dblp.org}
}

@article{prbench,
  author       = {Afra Feyza Aky{\"{u}}rek and
                  Advait Gosai and
                  Chen Bo Calvin Zhang and
                  Vipul Gupta and
                  Jaehwan Jeong and
                  Anisha Gunjal and
                  Tahseen Rabbani and
                  Maria Mazzone and
                  David Randolph and
                  Mohammad Mahmoudi Meymand and
                  Gurshaan Chattha and
                  Paula Rodriguez and
                  Diego Mares and
                  Pavit Singh and
                  Michael Liu and
                  Subodh Chawla and
                  Pete Cline and
                  Lucy Ogaz and
                  Ernesto Hernandez and
                  Zihao Wang and
                  Pavi Bhatter and
                  Marcos Ayestaran and
                  Bing Liu and
                  Yunzhong He},
  title        = {PRBench: Large-Scale Expert Rubrics for Evaluating High-Stakes Professional
                  Reasoning},
  journal      = {CoRR},
  volume       = {abs/2511.11562},
  year         = {2025},
  url          = {https://doi.org/10.48550/arXiv.2511.11562},
  doi          = {10.48550/ARXIV.2511.11562},
  eprinttype    = {arXiv},
  eprint       = {2511.11562},
  bibsource    = {dblp computer science bibliography, https://dblp.org}
}

@inproceedings{carmo,
  author       = {Taneesh Gupta and
                  Shivam Shandilya and
                  Xuchao Zhang and
                  Rahul Madhavan and
                  Supriyo Ghosh and
                  Chetan Bansal and
                  Huaxiu Yao and
                  Saravan Rajmohan},
  editor       = {Wanxiang Che and
                  Joyce Nabende and
                  Ekaterina Shutova and
                  Mohammad Taher Pilehvar},
  title        = {{CARMO:} Dynamic Criteria Generation for Context Aware Reward Modelling},
  booktitle    = {Findings of the Association for Computational Linguistics, {ACL} 2025,
                  Vienna, Austria, July 27 - August 1, 2025},
  series       = {Findings of {ACL}},
  volume       = {{ACL} 2025},
  pages        = {2202--2261},
  publisher    = {Association for Computational Linguistics},
  year         = {2025},
  url          = {https://aclanthology.org/2025.findings-acl.114/},
  bibsource    = {dblp computer science bibliography, https://dblp.org}
}

@article{rlrr,
  author       = {Anisha Gunjal and
                  Anthony Wang and
                  Elaine Lau and
                  Vaskar Nath and
                  Bing Liu and
                  Sean Hendryx},
  title        = {Rubrics as Rewards: Reinforcement Learning Beyond Verifiable Domains},
  journal      = {CoRR},
  volume       = {abs/2507.17746},
  year         = {2025},
  url          = {https://doi.org/10.48550/arXiv.2507.17746},
  doi          = {10.48550/ARXIV.2507.17746},
  eprinttype    = {arXiv},
  eprint       = {2507.17746},
  bibsource    = {dblp computer science bibliography, https://dblp.org}
}

@article{rlcf,
  author       = {Vijay Viswanathan and
                  Yanchao Sun and
                  Shuang Ma and
                  Xiang Kong and
                  Meng Cao and
                  Graham Neubig and
                  Tongshuang Wu},
  title        = {Checklists Are Better Than Reward Models For Aligning Language Models},
  journal      = {CoRR},
  volume       = {abs/2507.18624},
  year         = {2025},
  url          = {https://doi.org/10.48550/arXiv.2507.18624},
  doi          = {10.48550/ARXIV.2507.18624},
  eprinttype    = {arXiv},
  eprint       = {2507.18624},
  bibsource    = {dblp computer science bibliography, https://dblp.org}
}

@article{rubicon,
  author       = {Zenan Huang and
                  Yihong Zhuang and
                  Guoshan Lu and
                  Zeyu Qin and
                  Haokai Xu and
                  Tianyu Zhao and
                  Ru Peng and
                  Jiaqi Hu and
                  Zhanming Shen and
                  Xiaomeng Hu and
                  Xijun Gu and
                  Peiyi Tu and
                  Jiaxin Liu and
                  Wenyu Chen and
                  Yuzhuo Fu and
                  Zhiting Fan and
                  Yanmei Gu and
                  Yuanyuan Wang and
                  Zhengkai Yang and
                  Jianguo Li and
                  Junbo Zhao},
  title        = {Reinforcement Learning with Rubric Anchors},
  journal      = {CoRR},
  volume       = {abs/2508.12790},
  year         = {2025},
  url          = {https://doi.org/10.48550/arXiv.2508.12790},
  doi          = {10.48550/ARXIV.2508.12790},
  eprinttype    = {arXiv},
  eprint       = {2508.12790},
  bibsource    = {dblp computer science bibliography, https://dblp.org}
}

@article{ruscarl,
  author       = {Yang Zhou and
                  Sunzhu Li and
                  Shunyu Liu and
                  Wenkai Fang and
                  Jiale Zhao and
                  Jingwen Yang and
                  Jianwei Lv and
                  Kongcheng Zhang and
                  Yihe Zhou and
                  Hengtong Lu and
                  Wei Chen and
                  Yan Xie and
                  Mingli Song},
  title        = {Breaking the Exploration Bottleneck: Rubric-Scaffolded Reinforcement
                  Learning for General {LLM} Reasoning},
  journal      = {CoRR},
  volume       = {abs/2508.16949},
  year         = {2025},
  url          = {https://doi.org/10.48550/arXiv.2508.16949},
  doi          = {10.48550/ARXIV.2508.16949},
  eprinttype    = {arXiv},
  eprint       = {2508.16949},
  bibsource    = {dblp computer science bibliography, https://dblp.org}
}

@article{openrubrics,
  author       = {Tianci Liu and
                  Ran Xu and
                  Tony Yu and
                  Ilgee Hong and
                  Carl Yang and
                  Tuo Zhao and
                  Haoyu Wang},
  title        = {OpenRubrics: Towards Scalable Synthetic Rubric Generation for Reward
                  Modeling and {LLM} Alignment},
  journal      = {CoRR},
  volume       = {abs/2510.07743},
  year         = {2025},
  url          = {https://doi.org/10.48550/arXiv.2510.07743},
  doi          = {10.48550/ARXIV.2510.07743},
  eprinttype    = {arXiv},
  eprint       = {2510.07743},
  bibsource    = {dblp computer science bibliography, https://dblp.org}
}

@article{online-rubrics,
  author       = {MohammadHossein Rezaei and
                  Robert Vacareanu and
                  Zihao Wang and
                  Clinton Wang and
                  Bing Liu and
                  Yunzhong He and
                  Afra Feyza Aky{\"{u}}rek},
  title        = {Online Rubrics Elicitation from Pairwise Comparisons},
  journal      = {CoRR},
  volume       = {abs/2510.07284},
  year         = {2025},
  url          = {https://doi.org/10.48550/arXiv.2510.07284},
  doi          = {10.48550/ARXIV.2510.07284},
  eprinttype    = {arXiv},
  eprint       = {2510.07284},
  bibsource    = {dblp computer science bibliography, https://dblp.org}
}

@article{auto-rubrics,
  author       = {Lipeng Xie and
                  Sen Huang and
                  Zhuo Zhang and
                  Anni Zou and
                  Yunpeng Zhai and
                  Dingchao Ren and
                  Kezun Zhang and
                  Haoyuan Hu and
                  Boyin Liu and
                  Haoran Chen and
                  Zhaoyang Liu and
                  Bolin Ding},
  title        = {Auto-Rubric: Learning to Extract Generalizable Criteria for Reward
                  Modeling},
  journal      = {CoRR},
  volume       = {abs/2510.17314},
  year         = {2025},
  url          = {https://doi.org/10.48550/arXiv.2510.17314},
  doi          = {10.48550/ARXIV.2510.17314},
  eprinttype    = {arXiv},
  eprint       = {2510.17314},
  bibsource    = {dblp computer science bibliography, https://dblp.org}
}

@article{rubric-hub,
  author       = {Sunzhu Li and
                  Jiale Zhao and
                  Miteto Wei and
                  Huimin Ren and
                  Yang Zhou and
                  Jingwen Yang and
                  Shunyu Liu and
                  Kaike Zhang and
                  Wei Chen},
  title        = {RubricHub: {A} Comprehensive and Highly Discriminative Rubric Dataset
                  via Automated Coarse-to-Fine Generation},
  journal      = {CoRR},
  volume       = {abs/2601.08430},
  year         = {2026},
  url          = {https://doi.org/10.48550/arXiv.2601.08430},
  doi          = {10.48550/ARXIV.2601.08430},
  eprinttype    = {arXiv},
  eprint       = {2601.08430},
  bibsource    = {dblp computer science bibliography, https://dblp.org}
}

@article{rubric-arm,
  author       = {Ran Xu and Tianci Liu and Zihan Dong and Tony Yu and Ilgee Hong and Carl Yang and Linjun Zhang and Tao Zhao and Haoyu Wang},
  title        = {Alternating Reinforcement Learning for Rubric-Based Reward Modeling in Non-Verifiable LLM Post-Training},
  journal      = {CoRR},
  volume       = {abs/2602.01511},
  year         = {2026},
  url          = {https://arxiv.org/abs/2602.01511},
  eprinttype    = {arXiv},
  eprint       = {2602.01511}
}

@inproceedings{judge-anything,
  author       = {Shu Pu and
                  Yaochen Wang and
                  Dongping Chen and
                  Yuhang Chen and
                  Guohao Wang and
                  Qi Qin and
                  Zhongyi Zhang and
                  Zhiyuan Zhang and
                  Zetong Zhou and
                  Shuang Gong and
                  Yi Gui and
                  Yao Wan and
                  Philip S. Yu},
  editor       = {Luiza Antonie and
                  Jian Pei and
                  Xiaohui Yu and
                  Flavio Chierichetti and
                  Hady W. Lauw and
                  Yizhou Sun and
                  Srinivasan Parthasarathy},
  title        = {Judge Anything: {MLLM} as a Judge Across Any Modality},
  booktitle    = {Proceedings of the 31st {ACM} {SIGKDD} Conference on Knowledge Discovery
                  and Data Mining, V.2, {KDD} 2025, Toronto ON, Canada, August 3-7,
                  2025},
  pages        = {5742--5753},
  publisher    = {{ACM}},
  year         = {2025},
  url          = {https://doi.org/10.1145/3711896.3737409},
  doi          = {10.1145/3711896.3737409},
  bibsource    = {dblp computer science bibliography, https://dblp.org}
}

@article{multi-crit,
  author       = {Tianyi Xiong and
                  Yi Ge and
                  Ming Li and
                  Zuolong Zhang and
                  Pranav Kulkarni and
                  Kaishen Wang and
                  Qi He and
                  Zeying Zhu and
                  Chenxi Liu and
                  Ruibo Chen and
                  Tong Zheng and
                  Yanshuo Chen and
                  Xiyao Wang and
                  Renrui Zhang and
                  Wenhu Chen and
                  Heng Huang},
  title        = {Multi-Crit: Benchmarking Multimodal Judges on Pluralistic Criteria-Following},
  journal      = {CoRR},
  volume       = {abs/2511.21662},
  year         = {2025},
  url          = {https://doi.org/10.48550/arXiv.2511.21662},
  doi          = {10.48550/ARXIV.2511.21662},
  eprinttype    = {arXiv},
  eprint       = {2511.21662},
  bibsource    = {dblp computer science bibliography, https://dblp.org}
}

@article{omni-rrm,
  author       = {Zicheng Kong and Dehua Ma and Zhenbo Xu and Alven Yang and Yiwei Ru and Haoran Wang and Zixuan Zhou and Fuqing Bie and Liuyu Xiang and Huijia Wu and Jian Zhao and Zhaofeng He},
  title        = {Omni-RRM: Advancing Omni Reward Modeling via Automatic Rubric-Grounded Preference Synthesis},
  journal      = {CoRR},
  volume       = {abs/2602.00846},
  year         = {2026},
  url          = {https://arxiv.org/abs/2602.00846},
  eprinttype    = {arXiv},
  eprint       = {2602.00846},
}

@inproceedings{mllm-as-a-judge,
  author       = {Dongping Chen and
                  Ruoxi Chen and
                  Shilin Zhang and
                  Yaochen Wang and
                  Yinuo Liu and
                  Huichi Zhou and
                  Qihui Zhang and
                  Yao Wan and
                  Pan Zhou and
                  Lichao Sun},
  editor       = {Ruslan Salakhutdinov and
                  Zico Kolter and
                  Katherine A. Heller and
                  Adrian Weller and
                  Nuria Oliver and
                  Jonathan Scarlett and
                  Felix Berkenkamp},
  title        = {MLLM-as-a-Judge: Assessing Multimodal LLM-as-a-Judge with Vision-Language
                  Benchmark},
  booktitle    = {Forty-first International Conference on Machine Learning, {ICML} 2024,
                  Vienna, Austria, July 21-27, 2024},
  series       = {Proceedings of Machine Learning Research},
  volume       = {235},
  pages        = {6562--6595},
  publisher    = {{PMLR} / OpenReview.net},
  year         = {2024},
  url          = {https://proceedings.mlr.press/v235/chen24h.html},
  bibsource    = {dblp computer science bibliography, https://dblp.org}
}

@inproceedings{vl-rewardbench,
  author       = {Lei Li and
                  Yuancheng Wei and
                  Zhihui Xie and
                  Xuqing Yang and
                  Yifan Song and
                  Peiyi Wang and
                  Chenxin An and
                  Tianyu Liu and
                  Sujian Li and
                  Bill Yuchen Lin and
                  Lingpeng Kong and
                  Qi Liu},
  title        = {VL-RewardBench: {A} Challenging Benchmark for Vision-Language Generative
                  Reward Models},
  booktitle    = {{IEEE/CVF} Conference on Computer Vision and Pattern Recognition,
                  {CVPR} 2025, Nashville, TN, USA, June 11-15, 2025},
  pages        = {24657--24668},
  publisher    = {Computer Vision Foundation / {IEEE}},
  year         = {2025},
  url          = {https://openaccess.thecvf.com/content/CVPR2025/html/Li\_VL-RewardBench\_A\_Challenging\_Benchmark\_for\_Vision-Language\_Generative\_Reward\_Models\_CVPR\_2025\_paper.html},
  doi          = {10.1109/CVPR52734.2025.02296},
  bibsource    = {dblp computer science bibliography, https://dblp.org}
}

@article{mm-rewardbench,
  author       = {Michihiro Yasunaga and
                  Luke Zettlemoyer and
                  Marjan Ghazvininejad},
  title        = {Multimodal RewardBench: Holistic Evaluation of Reward Models for Vision
                  Language Models},
  journal      = {CoRR},
  volume       = {abs/2502.14191},
  year         = {2025},
  url          = {https://doi.org/10.48550/arXiv.2502.14191},
  doi          = {10.48550/ARXIV.2502.14191},
  eprinttype    = {arXiv},
  eprint       = {2502.14191},
  bibsource    = {dblp computer science bibliography, https://dblp.org}
}

@inproceedings{visionarena,
  author       = {Christopher Chou and
                  Lisa Dunlap and
                  Koki Mashita and
                  Krishna Mandal and
                  Trevor Darrell and
                  Ion Stoica and
                  Joseph E. Gonzalez and
                  Wei{-}Lin Chiang},
  title        = {VisionArena: 230k Real World User-VLM Conversations with Preference
                  Labels},
  booktitle    = {{IEEE/CVF} Conference on Computer Vision and Pattern Recognition,
                  {CVPR} 2025, Nashville, TN, USA, June 11-15, 2025},
  pages        = {3877--3887},
  publisher    = {Computer Vision Foundation / {IEEE}},
  year         = {2025},
  url          = {https://openaccess.thecvf.com/content/CVPR2025/html/Chou\_VisionArena\_230k\_Real\_World\_User-VLM\_Conversations\_with\_Preference\_Labels\_CVPR\_2025\_paper.html},
  doi          = {10.1109/CVPR52734.2025.00367},
  bibsource    = {dblp computer science bibliography, https://dblp.org}
}

@misc{gpt-5.4,
  author       = {OpenAI},
  title        = {GPT-5.4 Thinking System Card},
  howpublished = {\url{https://openai.com/index/gpt-5-4-thinking-system-card/}},
  year         = {2026},
  note         = {Released March 5, 2026}
}

@article{gemini3,
  title   = {Gemini 3: A Family of Highly Capable Multimodal Reasoning Models},
  author  = {Gemini Team, Google},
  journal = {arXiv preprint arXiv:2512.03267},
  year    = {2025},
  url     = {https://arxiv.org/abs/2512.03267}
}

@misc{claude-opus-4.6,
  author       = {Anthropic},
  title        = {Introducing Claude Opus 4.6},
  howpublished = {\url{https://www.anthropic.com/news/introducing-claude-opus-4-6}},
  year         = {2026},
  note         = {Released February 5, 2026}
}

@article{grpo,
  author       = {Zhihong Shao and
                  Peiyi Wang and
                  Qihao Zhu and
                  Runxin Xu and
                  Junxiao Song and
                  Mingchuan Zhang and
                  Y. K. Li and
                  Y. Wu and
                  Daya Guo},
  title        = {DeepSeekMath: Pushing the Limits of Mathematical Reasoning in Open
                  Language Models},
  journal      = {CoRR},
  volume       = {abs/2402.03300},
  year         = {2024},
  url          = {https://doi.org/10.48550/arXiv.2402.03300},
  doi          = {10.48550/ARXIV.2402.03300},
  eprinttype    = {arXiv},
  eprint       = {2402.03300},
  bibsource    = {dblp computer science bibliography, https://dblp.org}
}

@inproceedings{ai2d,
  author       = {Aniruddha Kembhavi and
                  Mike Salvato and
                  Eric Kolve and
                  Min Joon Seo and
                  Hannaneh Hajishirzi and
                  Ali Farhadi},
  editor       = {Bastian Leibe and
                  Jiri Matas and
                  Nicu Sebe and
                  Max Welling},
  title        = {A Diagram is Worth a Dozen Images},
  booktitle    = {Computer Vision - {ECCV} 2016 - 14th European Conference, Amsterdam,
                  The Netherlands, October 11-14, 2016, Proceedings, Part {IV}},
  series       = {Lecture Notes in Computer Science},
  pages        = {235--251},
  publisher    = {Springer},
  year         = {2016},
  url          = {https://doi.org/10.1007/978-3-319-46493-0\_15},
  doi          = {10.1007/978-3-319-46493-0\_15},
  bibsource    = {dblp computer science bibliography, https://dblp.org}
}

@inproceedings{chartqa,
  author       = {Ahmed Masry and
                  Do Xuan Long and
                  Jia Qing Tan and
                  Shafiq R. Joty and
                  Enamul Hoque},
  editor       = {Smaranda Muresan and
                  Preslav Nakov and
                  Aline Villavicencio},
  title        = {ChartQA: {A} Benchmark for Question Answering about Charts with Visual
                  and Logical Reasoning},
  booktitle    = {Findings of the Association for Computational Linguistics: {ACL} 2022,
                  Dublin, Ireland, May 22-27, 2022},
  series       = {Findings of {ACL}},
  pages        = {2263--2279},
  publisher    = {Association for Computational Linguistics},
  year         = {2022},
  url          = {https://doi.org/10.18653/v1/2022.findings-acl.177},
  doi          = {10.18653/V1/2022.FINDINGS-ACL.177},
  bibsource    = {dblp computer science bibliography, https://dblp.org}
}

@inproceedings{m3cot,
  author       = {Qiguang Chen and
                  Libo Qin and
                  Jin Zhang and
                  Zhi Chen and
                  Xiao Xu and
                  Wanxiang Che},
  editor       = {Lun{-}Wei Ku and
                  Andre Martins and
                  Vivek Srikumar},
  title        = {M{\({^3}\)}CoT: {A} Novel Benchmark for Multi-Domain Multi-step Multi-modal
                  Chain-of-Thought},
  booktitle    = {Proceedings of the 62nd Annual Meeting of the Association for Computational
                  Linguistics (Volume 1: Long Papers), {ACL} 2024, Bangkok, Thailand,
                  August 11-16, 2024},
  pages        = {8199--8221},
  publisher    = {Association for Computational Linguistics},
  year         = {2024},
  url          = {https://doi.org/10.18653/v1/2024.acl-long.446},
  doi          = {10.18653/V1/2024.ACL-LONG.446},
  bibsource    = {dblp computer science bibliography, https://dblp.org}
}


\newpage

\appendix

\section{Prompt Templates for Rubric Construction and Scoring}
\label{sec:prompt}

This appendix presents the complete system prompts used across the stages of rubric generation, rubric aggregation, and reward modeling, alongside the JSON schemas for their respective outputs. Each prompt is delivered as the system message to the reasoning model; the corresponding user message supplies the image and question, together with any stage-specific auxiliary inputs (ground-truth annotations or candidate rubric lists).

\subsection{Output JSON Schemas}
\label{ssec:json_schema}

Both rubric generation and response scoring produce structured JSON outputs, which facilitate reliable parsing and enable fine-grained programmatic aggregation and filtering.

\paragraph{Rubric Schema.}
The checklist-style rubric is a JSON object with two top-level arrays: \textit{essential} and \textit{additional}, corresponding to the two criterion categories defined in Section~\ref{ssec:prep_rubric}.
Each element in either array is a triplet:
\begin{itemize}[topsep=2pt, itemsep=1pt]
  \item \texttt{criterion}: a concrete, observable assertion that the judge can verify directly from the image, question, and candidate response.
  \item \texttt{reference}: the ground-truth answer strictly grounded in image facts or common knowledge.
  \item \texttt{weight}: a three-level integer encoding the criterion's importance---\textbf{1} (Auxiliary: supplementary information), \textbf{2} (Important: noticeably affects the user experience), \textbf{3} (Key: critical elements where any omission or deviation constitutes a definitive error).
\end{itemize}

\begin{tcolorbox}[
    colback=gray!5,
    colframe=gray!60,
    arc=3pt,
    boxrule=0.5pt,
    left=2mm, right=2mm, top=1.5mm, bottom=1.5mm
]
\begin{lstlisting}[basicstyle=\ttfamily\small, breaklines=true, showstringspaces=false]
{
  "essential": [
    {
      "criterion": "string",  // assertion to be verified
      "reference": "string",  // ground-truth
      "weight": 1 | 2 | 3     // importance
    }
  ]
  "additional": [
    ...
  ]
}
\end{lstlisting}
\end{tcolorbox}

\paragraph{Scoring Schema.}
The scoring output is a JSON array with one element per criterion, preserving the same ordering as the input rubric.
Each element contains three fields:
\begin{itemize}[topsep=2pt, itemsep=1pt]
  \item \texttt{criterion}: the assertion copied verbatim from the input rubric, serving as an explicit confirmation that the judge is addressing the correct item.
  \item \texttt{rationale}: a brief (1--2 sentence) explanation of the judgment, providing transparency and enabling post-hoc error analysis.
  \item \texttt{credit}: a three-level score---\textbf{0} (incorrect or missing), \textbf{0.5} (partially correct with minor errors), \textbf{1} (fully correct or semantically equivalent). Critically, \texttt{credit} is a \textit{qualitative correctness grade} independent of \texttt{weight}; the weighted aggregation across all scored criteria is performed externally (Eq.~\ref{eq:reward_agg}).
\end{itemize}

\begin{tcolorbox}[
    colback=gray!5,
    colframe=gray!60,
    arc=3pt,
    boxrule=0.5pt,
    left=2mm, right=2mm, top=1.5mm, bottom=1.5mm
]
\begin{lstlisting}[basicstyle=\ttfamily\small, breaklines=true, showstringspaces=false]
[
  {
    "criterion": "string",  // repeat the criterion to confirm
    "rationale": "string",  // rationale for the judgment
    "credit": 0 | 0.5 | 1   // assigned credit
  }
]
\end{lstlisting}
\end{tcolorbox}

\subsection{Rubric Generation}
\label{ssec:prompt_gen}

The generation prompt instructs the reasoning model to construct an instance-specific checklist-style rubric for a given image-question pair, following the principles described in Section~\ref{ssec:prep_rubric}:

\paragraph{Expert-Grounded Generation.}
When trustworthy ground-truth annotations are available, the prompt includes a \textit{dual verification} step: before finalizing the rubric, the model must confirm that the provided ground-truth fully satisfies all \textit{essential} criteria, ensuring that no correct response would be penalized by an erroneous check item.

\begin{tcolorbox}[
    enhanced, breakable,
    title={\small\textbf{System Prompt: Expert-Grounded Rubric Generation}},
    colback=gray!5,
    colframe=gray!60,
    colbacktitle=gray!22,
    coltitle=black,
    arc=3pt,
    boxrule=0.5pt,
    left=2mm, right=2mm, top=1.5mm, bottom=1.5mm,
    fonttitle=\small
]
\small
You are a multimodal evaluation expert. Given the user question and associated image, construct a instance-specific Checklist-Style Rubric for evaluating the accuracy of model responses. The criteria in this Checklist serve as the \textit{sole evaluation standard} for the reward model.

\medskip\noindent\textbf{Construction Principles}

\begin{itemize}[topsep=2pt, itemsep=0pt, leftmargin=1.8em]
  \item \textbf{Atomic:} Each check item targets exactly one key point or atomic sub-question within the query.
  \item \textbf{Comprehensive:} The combined set of items covers all critical dimensions of the user's question.
  \item \textbf{Precise:} Exclude redundant checks and checks unrelated to the question.
  \item \textbf{Objective:} Ground assessments in image facts or logical truths; avoid subjective uncertainty.
\end{itemize}

\medskip\noindent\textbf{Field Definitions}

\noindent Two categories of check items:
\begin{itemize}[topsep=2pt, itemsep=0pt, leftmargin=1.8em]
  \item \texttt{essential:} Core information prioritized by the query. Satisfying these is a prerequisite for a conceptually sound response regardless of its verbosity.
  \item \texttt{additional:} Relevant image facts, supplementary knowledge, or intermediate steps required to derive the answer. Optional, the list may be empty.
\end{itemize}

\noindent Fields per item:
\begin{itemize}[topsep=2pt, itemsep=0pt, leftmargin=1.8em]
  \item \texttt{criterion:} A concrete and observable assertion to be verified.
  \item \texttt{reference:} Ground-truth fact derived from the image and common knowledge.
  \item \texttt{weight:} A three-level integer quantifying the criterion's importance ranging from \texttt{1}~(Auxiliary: supplementary information) through \texttt{2}~(Important: noticeable impact on the user experience) to \texttt{3}~(Key: critical elements where any omission or deviation constitutes a definitive error).
\end{itemize}
\noindent Think from a scoring perspective: do not double-count and avoid over-decomposing key points.

\medskip\noindent\textbf{Dual Verification}

Before finalizing the Checklist, verify that the provided reference answer satisfies all \textit{essential} items. Any unsatisfied item indicates a construction error---the item is either incorrect or superfluous.

\smallskip\noindent\textbf{Output Format}

\{Structured JSON following the rubric schema in Appendix~\ref{ssec:json_schema}.\}
\end{tcolorbox}

\paragraph{Answer-Agnostic Generation.}
For datasets with noisy or absent annotations, the \textit{dual verification} step is omitted, preventing the model from anchoring on a potentially unreliable reference answer.

\begin{tcolorbox}[
    enhanced, breakable,
    title={\small\textbf{System Prompt: Answer-Agnostic Rubric Generation}},
    colback=gray!5,
    colframe=gray!60,
    colbacktitle=gray!22,
    coltitle=black,
    arc=3pt,
    boxrule=0.5pt,
    left=2mm, right=2mm, top=1.5mm, bottom=1.5mm,
    fonttitle=\small
]
\small
You are a multimodal evaluation expert. Given the user question and associated image, construct a instance-specific Checklist-Style Rubric for evaluating the accuracy of model responses. The criteria in this Checklist serve as the \textit{sole evaluation standard} for the reward model.

\medskip\noindent\textbf{Construction Principles}

\begin{itemize}[topsep=2pt, itemsep=0pt, leftmargin=1.8em]
  \item \textbf{Atomic:} Each check item targets exactly one key point or atomic sub-question within the query.
  \item \textbf{Comprehensive:} The combined set of items covers all critical dimensions of the user's question.
  \item \textbf{Precise:} Exclude redundant checks and checks unrelated to the question.
  \item \textbf{Objective:} Ground assessments in image facts or logical truths; avoid subjective uncertainty.
\end{itemize}

\medskip\noindent\textbf{Field Definitions}

\noindent Two categories of check items:
\begin{itemize}[topsep=2pt, itemsep=0pt, leftmargin=1.8em]
  \item \texttt{essential:} Core information prioritized by the query. Satisfying these is a prerequisite for a conceptually sound response regardless of its verbosity.
  \item \texttt{additional:} Relevant image facts, supplementary knowledge, or intermediate steps required to derive the answer. Optional, the list might be empty.
\end{itemize}

\noindent Fields per item:
\begin{itemize}[topsep=2pt, itemsep=0pt, leftmargin=1.8em]
  \item \texttt{criterion:} A concrete and observable assertion to be verified.
  \item \texttt{reference:} Ground-truth fact derived from the image and common knowledge.
  \item \texttt{weight:} A three-level integer quantifying the criterion's importance ranging from \texttt{1}~(Auxiliary: supplementary information) through \texttt{2}~(Important: noticeable impact on the user experience) to \texttt{3}~(Key: critical elements where any omission or deviation constitutes a definitive error).
\end{itemize}
\noindent Think from a scoring perspective: do not double-count and avoid over-decomposing key points.

\smallskip\noindent\textbf{Output Format}

\{Structured JSON following the rubric schema in Appendix~\ref{ssec:json_schema}.\}
\end{tcolorbox}

\subsection{Rubric Aggregation}
\label{ssec:prompt_agg}

After independent rubric generation by multiple models (Section~\ref{ssec:prep_rubric}), an aggregation prompt merges the candidate checklists into a single unified rubric. The model first applies the same four construction principles as in rubric generation, then executes the additional aggregation instructions below to apply majority-vote filtering, deduplicate overlapping check items, and verify the correctness of all references.

\begin{tcolorbox}[
    enhanced, breakable,
    title={\small\textbf{System Prompt: Rubric Aggregation}},
    colback=gray!5,
    colframe=gray!60,
    colbacktitle=gray!22,
    coltitle=black,
    arc=3pt,
    boxrule=0.5pt,
    left=2mm, right=2mm, top=1.5mm, bottom=1.5mm,
    fonttitle=\small
]
\small
You are a multimodal evaluation expert. Given the user question and associated image, construct a instance-specific Checklist-Style Rubric for evaluating the accuracy of model responses. The criteria in this Checklist serve as the \textit{sole evaluation standard} for the reward model.

\medskip\noindent\textbf{Construction Principles}

\begin{itemize}[topsep=2pt, itemsep=0pt, leftmargin=1.8em]
  \item \textbf{Atomic:} Each check item targets exactly one key point or atomic sub-question within the query.
  \item \textbf{Comprehensive:} The combined set of items covers all critical dimensions of the user's question.
  \item \textbf{Precise:} Exclude redundant checks and checks unrelated to the question.
  \item \textbf{Objective:} Ground assessments in image facts or logical truths; avoid subjective uncertainty.
\end{itemize}

\medskip\noindent\textbf{Field Definitions}

\noindent Two categories of check items:
\begin{itemize}[topsep=2pt, itemsep=0pt, leftmargin=1.8em]
  \item \texttt{essential:} Core information prioritized by the query. Satisfying these is a prerequisite for a conceptually sound response regardless of its verbosity.
  \item \texttt{additional:} Relevant image facts, supplementary knowledge, or intermediate steps required to derive the answer. Optional, the list might be empty.
\end{itemize}

\noindent Fields per item:
\begin{itemize}[topsep=2pt, itemsep=0pt, leftmargin=1.8em]
  \item \texttt{criterion:} A concrete and observable assertion to be verified.
  \item \texttt{reference:} Ground-truth fact derived from the image and common knowledge.
  \item \texttt{weight:} A three-level integer quantifying the criterion's importance ranging from \texttt{1}~(Auxiliary: supplementary information) through \texttt{2}~(Important: noticeable impact on the user experience) to \texttt{3}~(Key: critical elements where any omission or deviation constitutes a definitive error).
\end{itemize}
\noindent Think from a scoring perspective: do not double-count and avoid over-decomposing key points.

\medskip\noindent\textbf{Checklist Aggregation}

Do not construct the Checklist from scratch. Multiple models have each independently generated a candidate Checklist for the same question; your task is to merge them into a single unified rubric. Based on the construction principles, ensure that:
\begin{itemize}[topsep=2pt, itemsep=0pt, leftmargin=1.8em]
  \item All retained check items are necessary. Specifically:
    \begin{itemize}[topsep=1pt, itemsep=0pt, leftmargin=1.8em]
      \item Retain only items that the majority of candidates agree should be checked.
      \item Remove redundant or duplicate checks.
      \item Remove checks unrelated to the question.
    \end{itemize}
  \item Do not over-decompose check items.
  \item All \texttt{reference} values are correct.
\end{itemize}

\smallskip\noindent\textbf{Output Format}

\{Structured JSON following the rubric schema in Appendix~\ref{ssec:json_schema}.\}
\end{tcolorbox}

\subsection{Response Scoring}
\label{ssec:prompt_score}

The scoring prompt instructs the judge model to evaluate a candidate response against the finalized rubric, assigning a credit score to each criterion independently on a three-point scale.

\begin{tcolorbox}[
    enhanced, breakable,
    title={\small\textbf{System Prompt: Response Scoring}},
    colback=gray!5,
    colframe=gray!60,
    colbacktitle=gray!22,
    coltitle=black,
    arc=3pt,
    boxrule=0.5pt,
    left=2mm, right=2mm, top=1.5mm, bottom=1.5mm,
    fonttitle=\small
]
\small
You are a multimodal evaluation expert. Given the user question, associated image, and a Checklist-Style Rubric, evaluate the accuracy of the model response. Make sure that the criteria in the Checklist are used as the \textit{sole evaluation standard}.

\medskip\noindent\textbf{Checklist Fields}

\noindent Two categories of check items:
\begin{itemize}[topsep=2pt, itemsep=0pt, leftmargin=1.8em]
  \item \texttt{essential:} Core information prioritized by the query. Satisfying these is a prerequisite for a conceptually sound response regardless of its verbosity.
  \item \texttt{additional:} Relevant image facts, supplementary knowledge, or intermediate steps required to derive the answer. Optional, the list might be empty.
\end{itemize}

\noindent Fields per item:
\begin{itemize}[topsep=2pt, itemsep=0pt, leftmargin=1.8em]
  \item \texttt{criterion:} A concrete and observable assertion to be verified.
  \item \texttt{reference:} Ground-truth fact derived from the image and common knowledge.
\end{itemize}

\medskip\noindent\textbf{Scoring Fields}

For each criterion, output:
\begin{itemize}[topsep=2pt, itemsep=0pt, leftmargin=1.8em]
  \item \texttt{criterion:} The assertion to be verified (\textit{copied verbatim from input}).
  \item \texttt{rationale:} Core reasoning for the judgment (1--2 sentences).
  \item \texttt{credit:} A three-level score ranging from \texttt{0}~(No Credit: incorrect or missing) through \texttt{0.5}~(Partial Credit: partially correct with minor errors) to \texttt{1}~(Full Credit: fully correct or semantically equivalent).
\end{itemize}

\smallskip\noindent\textbf{Output Format}

\{Structured JSON following the scoring schema in Appendix~\ref{ssec:json_schema}.\}
\end{tcolorbox}
\newcommand{\caselabel}[2]{\noindent{\tiny\bfseries\color{#1}#2}}

\tcbset{
  caseRM/.style={
    enhanced,
    colback=white, colframe=blue!45!black,
    arc=4pt, boxrule=0.9pt,
    colbacktitle=blue!12, coltitle=blue!50!black,
    fonttitle=\small\bfseries,
    left=4mm, right=4mm, top=2mm, bottom=3mm,
  },
  casePP/.style={
    enhanced,
    colback=white, colframe=orange!70!black,
    arc=4pt, boxrule=0.9pt,
    colbacktitle=orange!12, coltitle=orange!60!black,
    fonttitle=\small\bfseries,
    left=4mm, right=4mm, top=2mm, bottom=3mm,
  },
  caseOP/.style={
    enhanced,
    colback=white, colframe=teal!60!black,
    arc=4pt, boxrule=0.9pt,
    colbacktitle=teal!12, coltitle=teal!50!black,
    fonttitle=\small\bfseries,
    left=4mm, right=4mm, top=2mm, bottom=3mm,
  },
}

\NewDocumentCommand{\CaseRM}{ m m +m +m +m m m +m }{
  \begin{tcolorbox}[caseRM, title={#1}]
  \noindent
  \begin{minipage}[t]{0.25\linewidth}
    \vspace{0pt}%
    \centering
    \includegraphics[width=\linewidth]{#2}
  \end{minipage}%
  \hfill%
  \begin{minipage}[t]{0.70\linewidth}
    \vspace{0pt}%
    \caselabel{blue!50!black}{Question:} \\
    \tiny\relax #3

    \medskip
    \caselabel{blue!50!black}{Response A:} \\
    \tiny\relax #4

    \medskip
    \caselabel{blue!50!black}{Response B:} \\
    \tiny\relax #5
  \end{minipage}
  \begin{xltabular}{\textwidth}{%
    >{\tiny}p{1.3cm}
    >{\tiny\hsize=#6\hsize\raggedright\arraybackslash}X
    >{\tiny\hsize=#7\hsize\raggedright\arraybackslash}X
  }
  \toprule
  \textbf{Type / Weight} & \textbf{Criterion / Reference} & \textbf{Rationale / Credit} \\
  \midrule
  \endhead
  #8
  \bottomrule
  \end{xltabular}
  \end{tcolorbox}
}

\NewDocumentCommand{\CaseOP}{ m m +m +m +m }{
  \begin{tcolorbox}[caseOP, title={#1}]
  \noindent
  \begin{minipage}[t]{0.25\linewidth}
    \vspace{0pt}%
    \centering
    \includegraphics[width=\linewidth]{#2}
  \end{minipage}%
  \hfill%
  \begin{minipage}[t]{0.70\linewidth}
    \vspace{0pt}%
    \caselabel{teal!50!black}{Question:} \\
    \tiny\relax #3

    \medskip
    \caselabel{red!50!black}{Before:} \\
    \tiny\relax #4

    \medskip
    \caselabel{green!50!black}{After:} \\
    \tiny\relax #5
  \end{minipage}
  \end{tcolorbox}
}

\section{Case Study of Reward Modeling}
\label{sec:case_reward}

\CaseRM{Count}{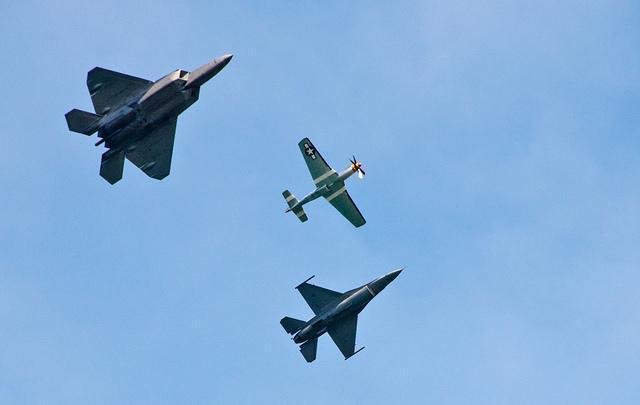}
{How many planes are visible?}
{Three planes are visible.}
{Three planes are visible. The one on top is an \textcolor{darkgreen}{F-22 stealth fighter}, the one in the middle is a \textcolor{darkgreen}{P-51 Mustang propeller plane}, and the one below is an \textcolor{darkgreen}{F-16 fighter jet}.}
{0.5}{1.5}{
Essential / 3 &
\textbf{Crit.:} Accurately state the total number of plane visible in the image. \par
\textbf{Ref.:} 3 &
\textbf{A:} The model response correctly identifies that there are three planes visible in the image, which matches the reference.~/~\textcolor{darkgreen}{\texttt{1}} \par
\textbf{B:} The model correctly states that there are three planes visible in the image, which matches the reference.~/~\textcolor{darkgreen}{\texttt{1}} \\
\midrule
Additional~/~1 &
\textbf{Crit.:} Identify the aircraft model in the top-left. \par
\textbf{Ref.:} F-22 Raptor &
\textbf{A:} The model response does not provide any identification of the aircraft models, so this criterion is not addressed.~/~\textcolor{red}{\texttt{0}} \par
\textbf{B:} The model identifies the top-left aircraft as an F-22 stealth fighter, which is correct and matches the reference.~/~\textcolor{darkgreen}{\texttt{1}} \\
\midrule
Additional~/~1 &
\textbf{Crit.:} Identify the aircraft model in the middle. \par
\textbf{Ref.:} P-51 Mustang &
\textbf{A:} The model response does not provide any identification of the aircraft models, so this criterion is not addressed.~/~\textcolor{red}{\texttt{0}} \par
\textbf{B:} The model correctly identifies the middle aircraft as a P-51 Mustang, which is accurate and aligns with the reference.~/~\textcolor{darkgreen}{\texttt{1}} \\
\midrule
Additional~/~1 &
\textbf{Crit.:} Identify the aircraft model at the bottom. \par
\textbf{Ref.:} F-16 Fighting Falcon &
\textbf{A:} The model response does not provide any identification of the aircraft models, so this criterion is not addressed.~/~\textcolor{red}{\texttt{0}} \par
\textbf{B:} The model correctly identifies the bottom aircraft as an F-16 fighter jet, which is accurate and matches the reference.~/~\textcolor{darkgreen}{\texttt{1}} \\
}

\newpage

\CaseRM{VQA}{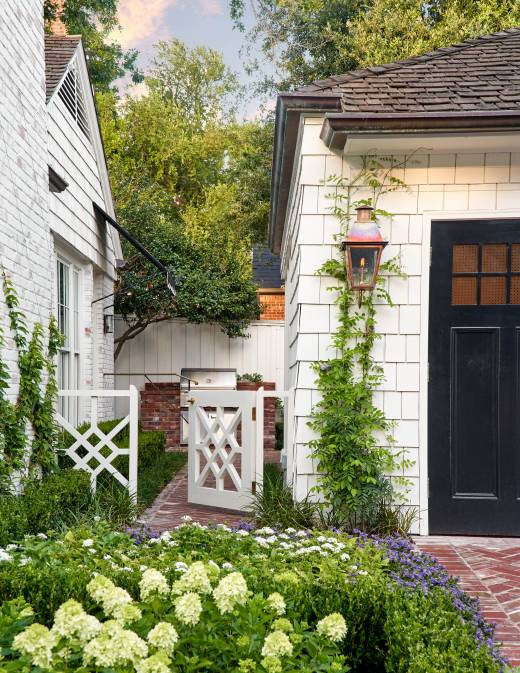}
{Is the door open?}
{Yes, the door is open. The image shows the white fence gate in an open position, providing access to the courtyard behind.}
{Yes, the gate in the center of the image is open. \textcolor{darkgreen}{The black door on the right is closed}.}
{0.5}{1.5}{
Essential / 3 &
\textbf{Crit.:} Determine the status of the white fence gate in the picture. \par
\textbf{Ref.:} The white fence gate in the picture is open. &
\textbf{A:} The model correctly identifies that the white fence gate is open, which aligns with the reference. The gate is visibly ajar, allowing access to the courtyard.~/~\textcolor{darkgreen}{\texttt{1}} \par
\textbf{B:} The model correctly identifies that the white gate in the center of the image is open, which matches the reference.~/~\textcolor{darkgreen}{\texttt{1}} \\
\midrule
Essential / 3 &
\textbf{Crit.:} Determine the status of the black door on the building on the right. \par
\textbf{Ref.:} The black door in the picture is closed. &
\textbf{A:} The model fails to address the black door on the right, which is closed. The response incorrectly implies the door is open, which is not supported by the image.~/~\textcolor{red}{\texttt{0}} \par
\textbf{B:} The model correctly states that the black door on the right is closed, which is consistent with the reference.~/~\textcolor{darkgreen}{\texttt{1}} \\
\midrule
Additional~/~1 &
\textbf{Crit.:} Determine the status of the door on the white building on the left. \par
\textbf{Ref.:} The door on the left in the picture is closed. &
\textbf{A:} The model does not mention the door on the left building, which is closed. While this is an additional criterion, the model's failure to address the primary open door (the gate) is the main issue.~/~\textcolor{red}{\texttt{0}} \par
\textbf{B:} The model does not address the status of the door on the left building, which is an additional criterion. However, since the question only asks about 'the door' and the most prominent door is the black one, the response is still considered accurate.~/~\textcolor{black}{\texttt{0.5}} \\
}

\medskip

\CaseRM{Knowledge}{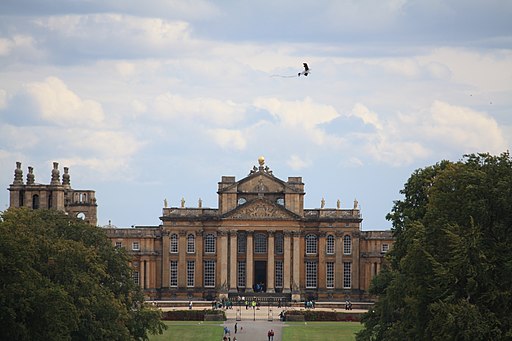}
{Which architect designed this historic house museum?}
{\textcolor{red}{Bridgeman} designed this historic house; he is a famous British master architect whose works integrate classical styles.}
{This historic house is \textcolor{darkgreen}{Blenheim Palace}, designed by English architect \textcolor{darkgreen}{Sir John Vanbrugh}. He incorporated \textcolor{darkgreen}{Baroque styles} to make the palace appear grand and magnificent.}
{0.6}{1.4}{
Essential / 3 &
\textbf{Crit.:} Accurately identify the name of the building in the picture. \par
\textbf{Ref.:} Blenheim Palace, also frequently called the Churchill Manor. &
\textbf{A:} The model response does not identify the building as Blenheim Palace, which is the correct name of the structure shown in the image. Instead, it incorrectly refers to 'Bridgeman,' which is not the name of the building.~/~\textcolor{red}{\texttt{0}} \par
\textbf{B:} The model correctly identifies the building as Blenheim Palace, which aligns with the reference.~/~\textcolor{darkgreen}{\texttt{1}} \\
\midrule
Essential~/~3 &
\textbf{Crit.:} Accurately state the name of the primary architect who designed the building. \par
\textbf{Ref.:} Sir John Vanbrugh. &
\textbf{A:} The model response incorrectly identifies the architect as 'Bridgeman,' which is factually incorrect. The actual primary architect of Blenheim Palace is Sir John Vanbrugh. The response fails to provide the correct name.~/~\textcolor{red}{\texttt{0}} \par
\textbf{B:} The model correctly states that Sir John Vanbrugh was the primary architect, matching the reference.~/~\textcolor{darkgreen}{\texttt{1}} \\
\midrule
Additional~/~1 &
\textbf{Crit.:} Mention another architect who made significant contributions to the design of the building. \par
\textbf{Ref.:} Nicholas Hawksmoor. &
\textbf{A:} The model response does not mention Nicholas Hawksmoor, who was a significant contributor to the design of Blenheim Palace. This information is missing from the response.~/~\textcolor{red}{\texttt{0}} \par
\textbf{B:} The model does not mention Nicholas Hawksmoor, who is noted in the reference as a significant contributor, so this criterion is not met.~/~\textcolor{red}{\texttt{0}} \\
\midrule
Additional~/~1 &
\textbf{Crit.:} Mention the architectural style of the building. \par
\textbf{Ref.:} English Baroque style. &
\textbf{A:} The model response does not mention the architectural style of the building, which is English Baroque. This information is absent from the response.~/~\textcolor{red}{\texttt{0}} \par
\textbf{B:} The model correctly identifies the architectural style as Baroque, which is consistent with the reference.~/~\textcolor{darkgreen}{\texttt{1}} \\
}

\newpage

\CaseRM{Molecular}{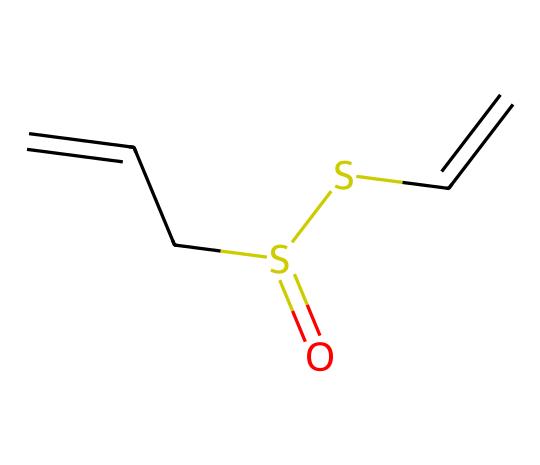}
{What is the molecular formula of allicin?}
{\textcolor{red}{The structure you provided is allicin}, and its molecular formula is \textcolor{red}{$C_6H_{10}S_2$}. It contains two sulfur atoms and two allyl groups, which is why it's also known as \textcolor{red}{diallyl disulfide}. It is the primary source of garlic's pungent flavor.}
{\textcolor{red}{Based on the image you provided}, the chemical formula of allicin (\textcolor{darkgreen}{chemical name: diallyl thiosulfinate}) is: \textcolor{darkgreen}{$C_6H_{10}OS_2$}.}
{0.9}{1.1}{
Essential / 3 &
\textbf{Crit.:} Correctly answer the text-based question and identify the molecular formula of Allicin. \par
\textbf{Ref.:} $C_6H_{10}OS_2$ &
\textbf{A:} The model incorrectly identifies the molecular formula as $C_6H_{10}S_2$, which is inaccurate. The correct molecular formula for allicin is $C_6H_{10}OS_2$. Additionally, the model misidentifies the compound in the image as allicin, which is incorrect.~/~\textcolor{red}{\texttt{0}} \par
\textbf{B:} The model correctly identifies the molecular formula of allicin as $C_6H_{10}OS_2$, which matches the reference.~/~\textcolor{darkgreen}{\texttt{1}} \\
\midrule
Additional / 2 &
\textbf{Crit.:} Clearly state that the molecule in the image is not the same substance as the Allicin mentioned in the question, and provide the correct molecular formula for the molecule in the image. \par
\textbf{Ref.:} The molecular structure in the image is different from that of allicin; its molecular formula is $C_5H_8OS_2$. &
\textbf{A:} The model fails to recognize that the molecule in the image is not allicin. It incorrectly states that the provided structure is allicin and does not provide the correct molecular formula for the actual molecule shown ($C_5H_8OS_2$).~/~\textcolor{red}{\texttt{0}} \par
\textbf{B:} The model fails to acknowledge that the molecule in the provided image is structurally different from allicin and does not provide the correct molecular formula ($C_5H_8OS_2$) for the structure shown.~/~\textcolor{red}{\texttt{0}} \\
\midrule
Additional~/~1 &
\textbf{Crit.:} Provide the chemical names for Allicin and the molecule in the image respectively. \par
\textbf{Ref.:} The chemical name of Allicin is diallyl thiosulfinate. The chemical name of the molecule in the image is S-allyl ethenesulfinothioate (or S-allyl vinyl thiosulfinate). &
\textbf{A:} The model does not provide the chemical name for allicin or the molecule in the image. It incorrectly identifies the image as allicin and calls it diallyl disulfide, which is a different compound.~/~\textcolor{red}{\texttt{0}} \par
\textbf{B:} The model provides the correct chemical name for allicin ($C_6H_{10}OS_2$) but does not provide the name for the molecule in the image.~/~\textcolor{black}{\texttt{0.5}} \\
\midrule
Additional~/~1 &
\textbf{Crit.:} Briefly explain the core structural difference between Allicin and the molecule in the image. \par
\textbf{Ref.:} Allicin's structure consists of two allyl groups attached to the central thiosulfinate group ($-S(O)S-$), whereas the molecule in the image consists of one allyl group and one vinyl group attached to that same central structure. &
\textbf{A:} The model does not explain any structural differences because it incorrectly assumes the image shows allicin. It fails to recognize the structural difference between allicin (two allyl groups) and the molecule in the image (one allyl and one vinyl group).~/~\textcolor{red}{\texttt{0}} \par
\textbf{B:} The model does not explain the structural difference between allicin and the molecule in the image.~/~\textcolor{red}{\texttt{0}} \\
}

\medskip

\CaseRM{TabMWP}{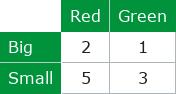}
{If Romero picks a marble at random, what is the probability that it is not red?}
{There are a total of 2 + 1 + 5 + 3 = 11 marbles, among which there are \textcolor{red}{2 + 1 = 3 red} ones. There are \textcolor{red}{11 - 3 = 8 non-red} marbles. Therefore, the probability of it being non-red is 8/11.}
{The total number of marbles is 2 + 1 + 5 + 3 = 11. The non-red ones are green, and there are \textcolor{darkgreen}{1 + 3 = 4}. Therefore, the probability of it being non-red is 4/11.}
{0.8}{1.2}{
Essential / 3 &
\textbf{Crit.:} Accurately provide the probability of randomly picking a non-red marble. \par
\textbf{Ref.:} 4/11 &
\textbf{A:} The model incorrectly calculates the number of red marbles as 3 (2 + 1) instead of 7 (2 + 5), leading to an incorrect count of non-red marbles (8 instead of 4) and a final probability of 8/11 instead of the correct 4/11.~/~\textcolor{red}{\texttt{0}} \par
\textbf{B:} The model correctly calculates the probability of picking a non-red marble as 4/11, which matches the reference value.~/~\textcolor{darkgreen}{\texttt{1}} \\
\midrule
Additional~/~1 &
\textbf{Crit.:} Correctly calculate the total number of all marbles. \par
\textbf{Ref.:} According to the table, the total number of marbles is $2 + 1 + 5 + 3 = 11$. &
\textbf{A:} The model correctly calculates the total number of marbles as 11 (2 + 1 + 5 + 3), which matches the reference.~/~\textcolor{darkgreen}{\texttt{1}} \par
\textbf{B:} The model correctly sums the values in the table (2 + 1 + 5 + 3) to find the total number of marbles as 11, which is accurate.~/~\textcolor{darkgreen}{\texttt{1}} \\
\midrule
Additional~/~1 &
\textbf{Crit.:} Correctly calculate the total number of non-red (i.e., green) marbles. \par
\textbf{Ref.:} According to the table, non-red marbles are the green marbles, and the count is $1 + 3 = 4$. &
\textbf{A:} The model incorrectly identifies non-red marbles as 8, but the correct count is 4 (1 green big + 3 green small). The model's error stems from misidentifying the red marbles.~/~\textcolor{red}{\texttt{0}} \par
\textbf{B:} The model correctly identifies the green marbles as the non-red ones and sums their counts (1 + 3) to get 4, which is accurate.~/~\textcolor{darkgreen}{\texttt{1}} \\
}

\newpage

\CaseRM{ChartQA}{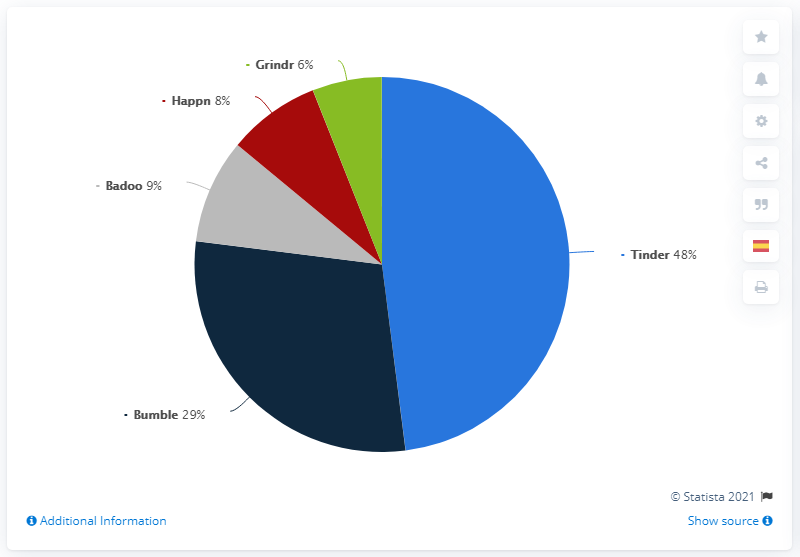}
{Which is the most popular dating app in Mexico in 2021?}
{In 2021, the most popular dating app in Mexico was Tinder, accounting for 48\% of the market. It was followed by \textcolor{darkgreen}{Bumble (29\%) and Badoo (9\%)}.}
{Based on the pie chart provided, the app with the largest share is Tinder at 48\%.

However, please note that the flag icon on the right side of the image is \textcolor{darkgreen}{the flag of Spain}, which suggests this data is for Spain, not Mexico.}
{0.8}{1.2}{
Essential / 3 &
\textbf{Crit.:} Based on the chart, clearly identify the most popular dating app and its market share. \par
\textbf{Ref.:} According to the chart, the most popular dating app is Tinder, with a market share of 48\%. &
\textbf{A:} The model correctly identifies Tinder as the most popular app with 48\% market share, which is directly supported by the chart data.~/~\textcolor{darkgreen}{\texttt{1}} \par
\textbf{B:} The model correctly identifies Tinder as the most popular app with a 48\% share, which is the largest segment in the pie chart.~/~\textcolor{darkgreen}{\texttt{1}} \\
\midrule
Additional~/~1 &
\textbf{Crit.:} Point out that the geographical scope of the chart data may not match the user's question and provide a reason. \par
\textbf{Ref.:} The user's question concerns Mexico, but there is a Spanish flag icon in the toolbar on the right side of the chart, which may indicate the data pertains to Spain rather than Mexico. &
\textbf{A:} The model fails to address the discrepancy between the question about Mexico and the Spanish flag icon in the chart, which suggests the data might be for Spain, not Mexico.~/~\textcolor{red}{\texttt{0}} \par
\textbf{B:} The model correctly observes the Spanish flag icon and correctly infers that the data is likely for Spain, not Mexico, which is a key point of discrepancy.~/~\textcolor{darkgreen}{\texttt{1}} \\
\midrule
Additional~/~1 &
\textbf{Crit.:} List the other dating apps in the chart and their shares in descending order of market share. \par
\textbf{Ref.:} The other apps in the chart and their market shares are: Bumble (29\%), Badoo (9\%), Happn (8\%), and Grindr (6\%). &
\textbf{A:} The model lists Bumble (29\%) and Badoo (9\%), which are correct, but omits the other apps (Happn, Grindr) and does not present them in descending order.~/~\textcolor{red}{\texttt{0}}.5 \par
\textbf{B:} The model does not list the other apps or their market shares, which is a missing piece of information from the checklist.~/~\textcolor{red}{\texttt{0}} \\
}

\medskip

\CaseRM{TQA}{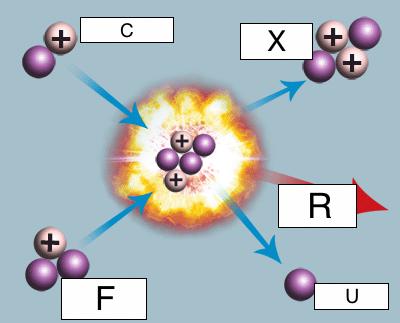}
{Which label refers to the neutron?}
{In the image, C is a positively charged particle, which is a proton. X is a neutral particle, uncharged, and is emitted from the nuclear reaction. Neutrons are uncharged and are common products of nuclear reactions. Therefore, X refers to the neutron.}
{The image shows the process of nuclear fusion, with atomic nuclei at the center surrounded by positively charged particles (protons) and neutral particles. Label U points to an uncharged particle, which fits the characteristics of a neutron.}
{1}{1}{
Essential / 3 &
\textbf{Crit.:} Correctly select the letter refers to the neutron. \par
\textbf{Ref.:} U &
\textbf{A:} The model incorrectly identifies label X as the neutron. According to the checklist, the correct label for the neutron is U. The model's answer is factually incorrect.~/~\textcolor{red}{\texttt{0}} \par
\textbf{B:} The model correctly identifies that label U refers to the neutron, which is the uncharged particle shown in the image.~/~\textcolor{darkgreen}{\texttt{1}} \\
\midrule
Additional / 2 &
\textbf{Crit.:} Explain why label U represents a neutron based on the image information. \par
\textbf{Ref.:} In the image, spheres with a "+" sign are protons, and the uncharged purple spheres are neutrons. Label U points to a single, uncharged purple sphere, therefore representing a neutron. &
\textbf{A:} The model does not address this criterion. It fails to explain why U represents a neutron, instead providing a flawed explanation for X. This criterion is completely missed.~/~\textcolor{red}{\texttt{0}} \par
\textbf{B:} The model correctly explains that U points to an uncharged particle, which is the defining characteristic of a neutron, consistent with the image's depiction.~/~\textcolor{darkgreen}{\texttt{1}} \\
\midrule
Additional~/~1 &
\textbf{Crit.:} Identify the physical process shown in the figure. \par
\textbf{Ref.:} The image shows a nuclear fusion reaction. Specifically, a deuterium nucleus (C) and a tritium nucleus (F) fuse to produce a helium nucleus (X) and a neutron (U), while releasing energy (R). &
\textbf{A:} The model does not identify the physical process shown in the figure. This criterion is not addressed at all.~/~\textcolor{red}{\texttt{0}} \par
\textbf{B:} The model correctly identifies the process as nuclear fusion, which is the correct interpretation of the image showing nuclei combining and releasing energy.~/~\textcolor{darkgreen}{\texttt{1}} \\
\midrule
Additional~/~1 &
\textbf{Crit.:} Explain why the other options (B, C, D) are incorrect by identifying the particles represented by labels X, F, and C. \par
\textbf{Ref.:} The other options are incorrect. C represents a deuterium nucleus (1 proton, 1 neutron), F represents a tritium nucleus (1 proton, 2 neutrons), and X represents a helium nucleus (2 protons, 2 neutrons). They are all atomic nuclei composed of multiple particles, rather than a single neutron. &
\textbf{A:} The model incorrectly identifies X as the neutron, so it cannot provide a correct explanation for why the other labels are incorrect. It fails to identify C and F correctly, and its explanation for X is wrong. This criterion is not addressed correctly.~/~\textcolor{red}{\texttt{0}} \par
\textbf{B:} The model does not address the other labels (X, F, C) or explain why they are incorrect, which is a significant omission from the checklist.~/~\textcolor{red}{\texttt{0}} \\
}

\newpage

\section{Case Study of Preference Optimization}
\label{sec:case_optim}

\CaseOP{Misleading}{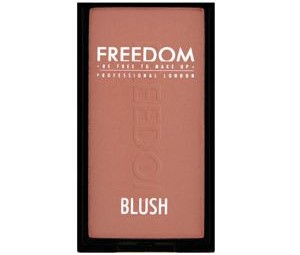}
{Based on the information on the cover, which series does this book likely belong to?}
{Based on the information on the cover, this book may belong to the "FREEDOM" series.}
{Based on the information in the picture, this is likely not a book, but a makeup product—a blush.

It belongs to the "FREEDOM" series, and the brand is Professional Formula. The packaging says "BE FREE TO MAKE UP," emphasizing the concept of makeup freedom. The shade of this blush is "BLUSH," a warm nude pink.}

\medskip

\CaseOP{Fine-Grained Perception}{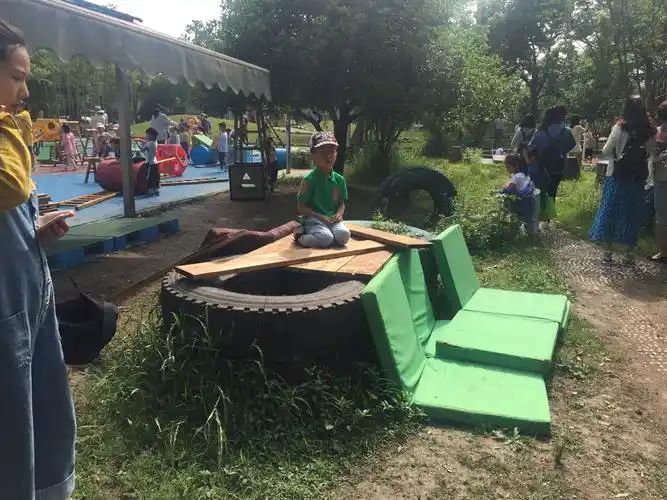}
{Is there a cobblestone road surface?}
{There's no cobblestone path in the picture; the ground is dirt and grass, with green mats nearby. It looks like an outdoor children's play area.}
{Yes, there is a cobblestone path on the right side of the picture. This path extends from the lower right corner of the image into the distance, with grassy areas and some tourists strolling along it.}

\medskip

\CaseOP{Knowledge}{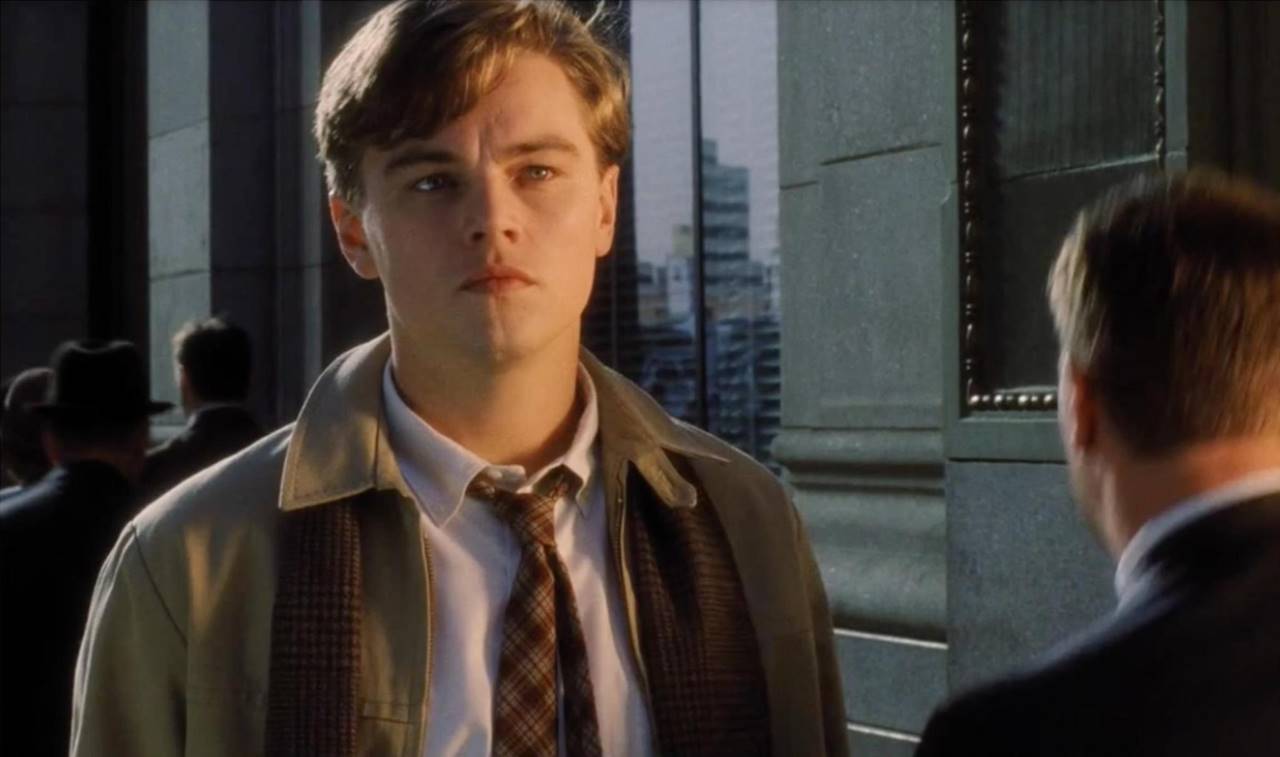}
{Which movie is this from?}
{This is a scene from the movie ``The Wolf of Wall Street'', in which Leonardo DiCaprio plays the protagonist, Jordan Belfort.}
{This is a scene from the film ``Catch Me If You Can'', starring Leonardo DiCaprio. Directed by Steven Spielberg and released in 2002, the film is an adaptation of Frank Abagnale's autobiography.}

\medskip

\CaseOP{Count}{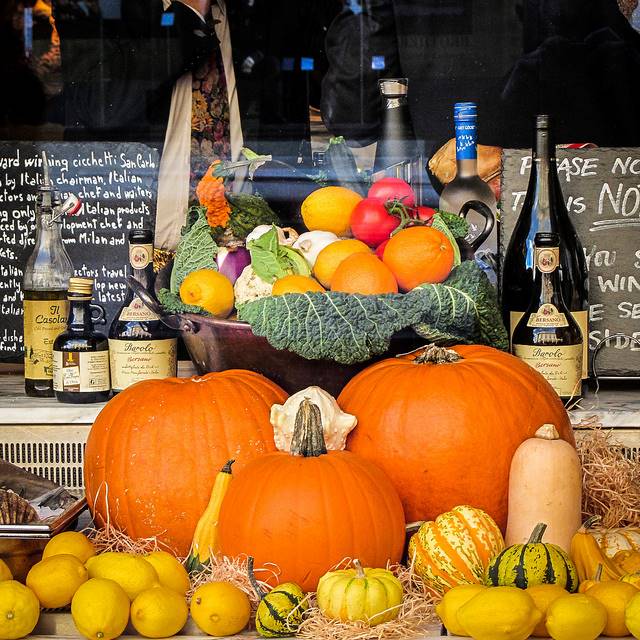}
{If the number of bottles doubles, how many bottles will you see?}
{There were originally 4 bottles; doubling that number results in 8.

Count the bottles in the picture: 3 on the left, 2 on the right, for a total of 5.

Therefore, if the number of bottles is doubled, you will see 10 bottles.}
{In the image, the following bottles are present:

- 3 small bottles on the left

- 1 bottle with blue cap in the middle

- 2 bottles on the right (one tall, one short)

- 1 bottle behind the fruit basket, partially visible

That's 7.

So if bottle count becomes 2 times, then 7 * 2 = 14.

Therefore, you would see 14 bottles.}

\medskip

\CaseOP{OCR}{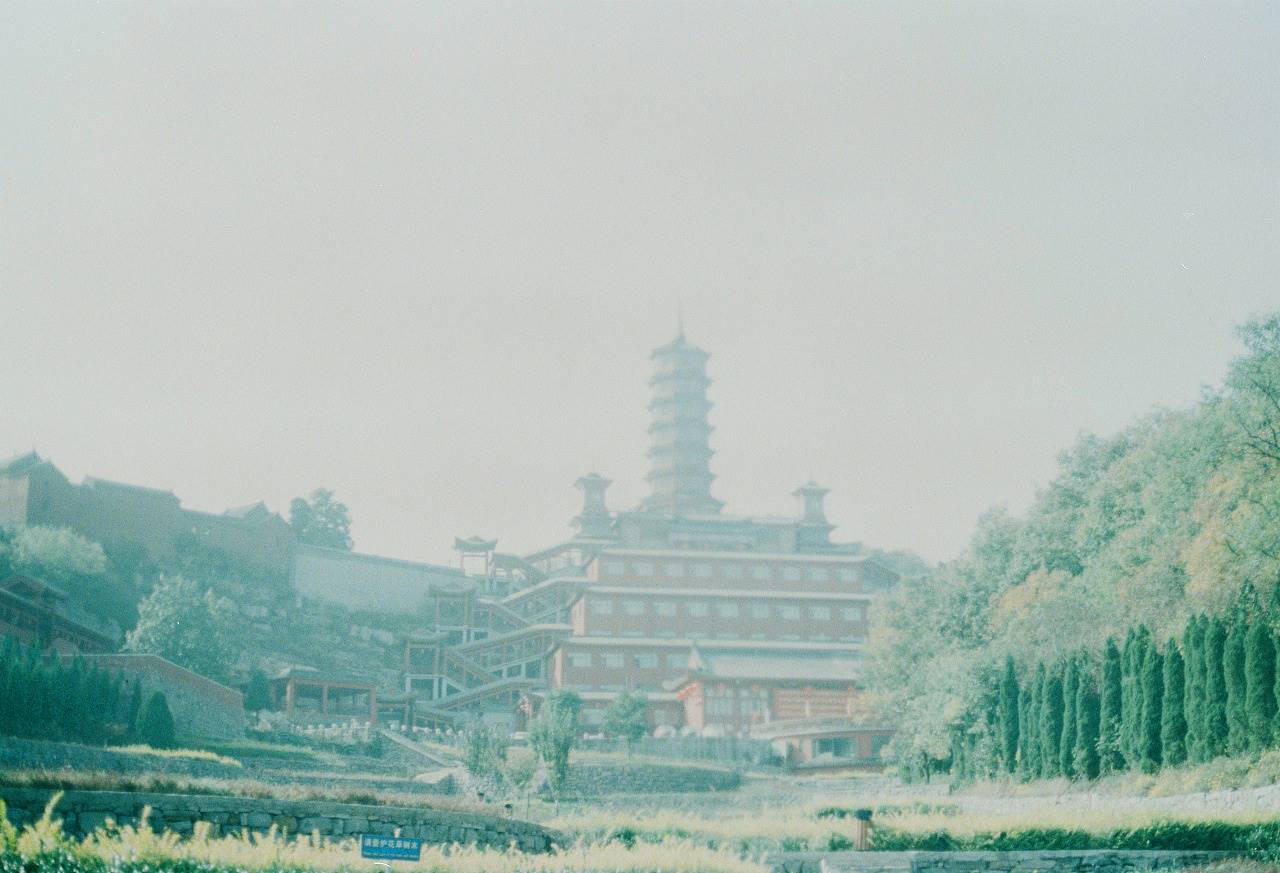}
{\zh{最下面的那个蓝色牌子写了啥}\\
What does the blue sign at the very bottom say?}
{\zh{牌子上写着“请勿在花丛中踩踏”。}\\
The sign says: "Do not trample on the flowers."}
{\zh{最下面的蓝色牌子上写着：“请爱护花草树木”}\\
The blue sign at the bottom says: "Please care for the flowers, grass, and trees."}

\medskip

\CaseOP{Implicit Reasoning}{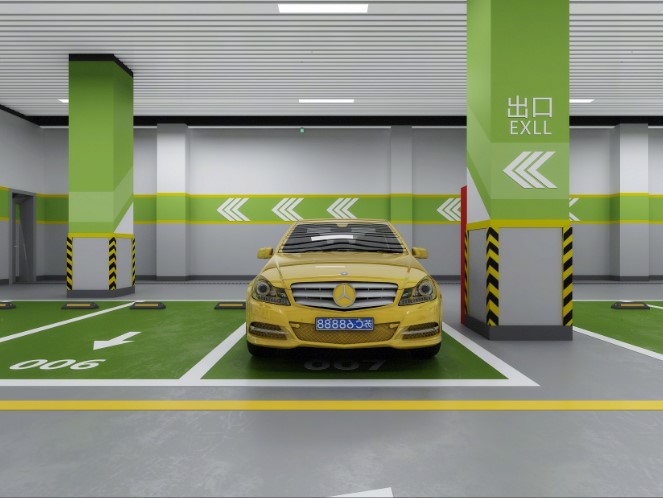}
{\zh{帮我记一下我的车位号}\\
Help me remember my parking space number.}
{\zh{好的，您的车位号是006。下次停车时记得找这个号码哦，我也会帮您记住的。}\\
Sure, your parking space number is 006. Remember to look for this number next time you park, and I will help you remember it as well.}
{\zh{好的，您的车位号是 007。旁边还有个 006 车位，您停在了绿色的标线区域内，位置很清晰。出口方向就在右边柱子上，写着“出口 EXLL”和箭头哦。}\\
Sure, your parking space number is 007. There is a space 006 next to it; you are parked within the green marked area, and the location is very clear. The exit direction is on the pillar to the right, which says "EXIT" with an arrow.}

\medskip

\CaseOP{Common Sense Reasoning}{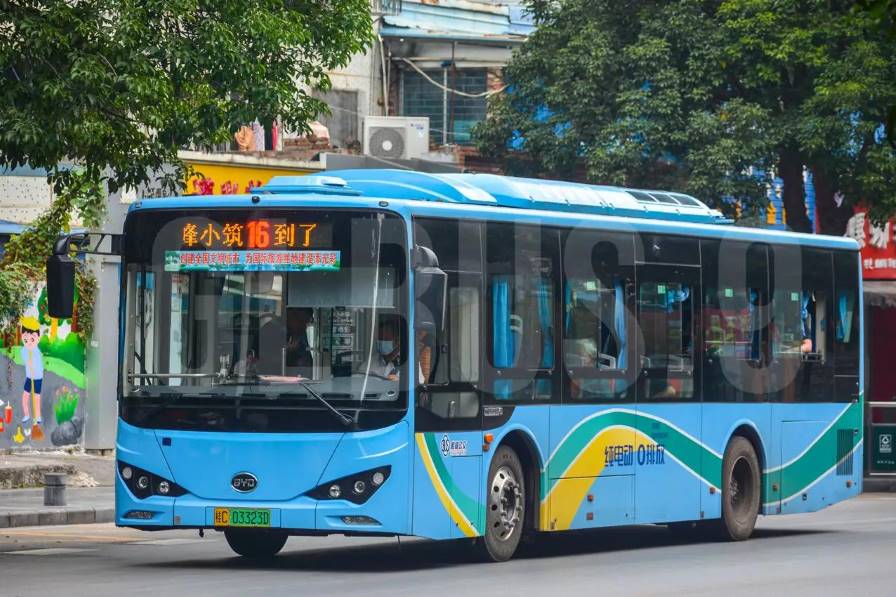}
{\zh{这是哪个城市的公交车？}\\
Which city is this bus from?}
{\zh{这辆公交车是广西柳州的。车身上有“柳州市公共交通集团”的标志，车牌是桂C开头，这是柳州的车牌代码。车头显示16路公交已到“犀小筑”站，车身写着“纯电动 0排放”，说明是当地的新能源公交车。}\\
This bus is from Liuzhou, Guangxi. The bus body features the "Liuzhou Public Transport Group" logo, and the license plate starts with "Gui C," which is the plate code for Liuzhou. The front of the bus shows that Route 16 has arrived at the "Xixiaozhu" station, and the body says "Pure Electric, 0 Emissions," indicating it is a local new energy bus.}
{\zh{这辆公交车是广西桂林的16路，车身显示“桂C·03323D”，属于桂林公交集团。车身上有“纯电动 0排放”的标识，说明是环保车型。从线路和车牌可以判断，这是桂林市的公交系统。}\\
This bus is Route 16 from Guilin, Guangxi. The body displays "Gui C·03323D" and belongs to the Guilin Bus Group. It has a "Pure Electric, 0 Emissions" label, indicating it is an environmentally friendly model. Based on the route and the license plate, it can be determined that this is part of the Guilin city bus system.}

\medskip

\CaseOP{Cuisine}{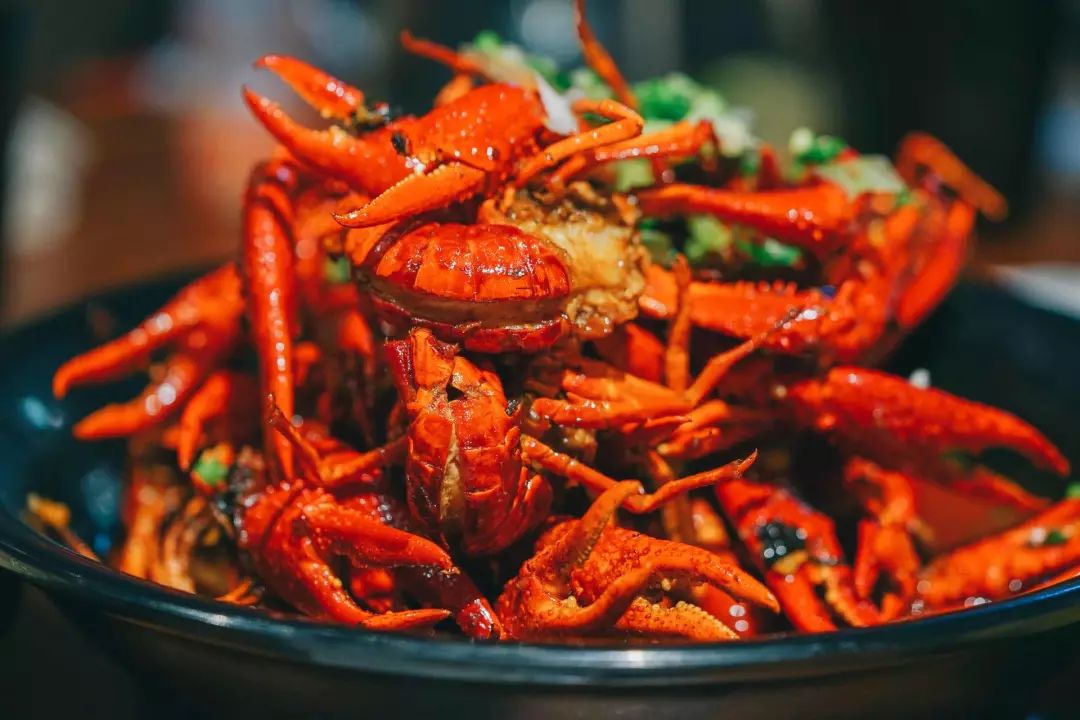}
{\zh{这个小龙虾开过背了吗}\\
Has this crawfish been deveined (back-cut)?}
{\zh{看这盘小龙虾，外壳红亮，个头饱满，确实挺诱人的。不过从图片上看，这些小龙虾的背部没有明显的开刀痕迹，应该还是完整的。不过有些店家会提前处理好，所以也有可能是特意没开背。您想吃的时候可以问问服务员哦。}\\
Looking at this plate of crawfish, the shells are bright red and the size is plump, which looks very tempting. however, from the image, there are no obvious cut marks on the backs of these crawfish; they appear to be intact. Some shops might process them beforehand, so it's possible they were intentionally not cut along the back. You can ask the waiter when you are ready to eat.}
{\zh{从图片来看，这些小龙虾的虾背是打开的。可以看到很多小龙虾的背部被切开，露出了里面的虾线和虾肉，这是烹饪前常见的处理方式，方便入味和食用。

另外，它们的颜色非常红亮，酱汁浓郁，看起来是经过了爆炒或香辣等烹饪手法处理过的。这种做法通常会先对小龙虾进行开背处理，再进行后续的烹饪步骤。所以可以判断，这些小龙虾已经开过背了。}\\
From the image, the backs of these crawfish are open. You can see that the backs of many crawfish have been cut, exposing the vein and the meat inside. This is a common processing method before cooking to make it easier to season and eat.
Additionally, their color is very bright red and the sauce is rich, appearing to have been stir-fried or cooked with spicy seasoning. This preparation method usually involves deveining the crawfish before the subsequent cooking steps. Therefore, it can be determined that these crawfish have been back-cut.}

\end{document}